\pdfoutput=1
\documentclass[11pt]{article}

\usepackage[preprint]{acl}

\usepackage{times}
\usepackage{latexsym}

\usepackage[T1]{fontenc}
\usepackage[utf8]{inputenc}

\usepackage{microtype}

\usepackage{inconsolata}

\usepackage{tabularx} 
\usepackage{hyperref}
\usepackage{url}
\usepackage{booktabs}
\usepackage{makecell}
\usepackage{graphicx}
\usepackage{algorithm}
\usepackage{amsmath, amssymb}
\usepackage{ragged2e}
\usepackage{enumitem}
\usepackage{algpseudocode}
\PassOptionsToPackage{most}{tcolorbox}
\usepackage{tcolorbox}
\usepackage{xcolor}
\usepackage{multirow}
\usepackage{lineno}
\definecolor{nmgray}{gray}{0.4}
\definecolor{darkorange}{rgb}{0.8, 0.4, 0.0}
\definecolor{deepblue}{rgb}{0.0, 0.0, 0.7}
\definecolor{deepgreen}{rgb}{0.0, 0.5, 0.0}
\definecolor{lightgreen}{RGB}{180, 216, 180}
\definecolor{lightblue}{rgb}{0.7, 0.75, 1.0}
\usepackage{colortbl}
\arrayrulecolor{black}

\usepackage{arydshln}
\usepackage{array}
\usepackage{adjustbox}
\usepackage{fontawesome}

\definecolor{darkblue}{rgb}{0, 0, 0.5}
\hypersetup{colorlinks=true, citecolor=darkblue, linkcolor=darkblue, urlcolor=darkblue}

\title{LogicTree: Structured Proof Exploration for Coherent and Rigorous Logical Reasoning with Large Language Models}

\author{Kang He \quad Kaushik Roy \\
         Electrical and Computer Engineering, Purdue University \\
         \texttt{\{he603, kaushik\}@purdue.edu}}

\begin{document}
\maketitle
\begin{abstract}
Large language models (LLMs) have achieved remarkable multi-step reasoning capabilities across various domains. 
However, LLMs still face distinct challenges in complex logical reasoning, as (1) proof-finding requires systematic exploration and the maintenance of logical coherence 
% throughout the process, 
and (2) searching the right combination of premises at each reasoning step is inherently challenging in tasks with large premise space. 
To address this, we propose LogicTree, an inference-time modular framework employing algorithm-guided search to automate structured proof exploration and ensure logical coherence. 
Advancing beyond tree-of-thought (ToT), we incorporate caching mechanism into LogicTree to enable effective utilization of historical knowledge,
% across branches
 preventing reasoning stagnation and minimizing redundancy. 
Furthermore, we address the combinatorial complexity of premise search by decomposing it into a linear process. The refined premise selection restricts subsequent inference to at most one derivation per step, enhancing reasoning granularity and enforcing strict step-by-step reasoning. 
Additionally, we introduce two LLM-free heuristics for premise prioritization, enabling strategic proof search.
%and improving proof search efficiency.
Experimental results on five datasets demonstrate that LogicTree optimally scales inference-time computation to achieve higher proof accuracy, surpassing chain-of-thought (CoT) and ToT with average gains of 23.6\% and 12.5\%, respectively, on GPT-4o. Moreover, within LogicTree, GPT-4o outperforms o3-mini by 7.6\% on average.\footnote{Our code is available at \url{https://github.com/kang-ml/LogicTree}.}
\end{abstract}

\section{Introduction}

Recent advances in large language models (LLMs), such as OpenAI’s o1/o3 series~\citep{openai2024o1, openai2025o3mini}, DeepSeek-R1~\citep{guo2025deepseek} and Grok-3~\citep{grok3}, have demonstrated remarkable reasoning capabilities in domains like code generation and complex mathematical problem-solving. 
% A key factor in their success is inference-time scaling~\citep{snell2024scaling}, which enables models to think deeper and generate long chain-of-thought (CoT) reasoning~\citep{wei2022chain, chen2025towards}.
However, \textit{logical reasoning} ~\citep{dowden2020logical, ijcai2020p0537} presents unique challenges that differentiate it from other reasoning domains \citep{liu2025logical, xu2025large}. It demands rigorous verification of a hypothesis through deliberate reasoning over a set of premises consisting of facts and rules, where two difficulties may arise. First, in complex problems, the precise proof path is not immediately apparent. Proof discovery requires systematic and extensive exploration ~\citep{saparov2023language}. Second, each reasoning step involves selecting relevant premises and inferring based on them. In a large premise space, difficulty in identifying the right fact-rule combination directly affects inference accuracy ~\citep{kazemi-etal-2023-lambada}.

To tackle these challenges, some studies use an iterative framework to build longer reasoning chains for solving complex problems ~\citep{creswell2023selectioninference}. Within the framework, they adopt a modular approach to decompose individual reasoning steps ~\citep{khot2023decomposed}, separating premise selection from inference and assigning each to specialized LLM modules for improved accuracy ~\citep{xu2024faithful, zhang2025cumulative, sun-etal-2024-determlr}. Further research integrates LLM modules into tree structures, enabling systematic proof exploration ~\citep{kazemi-etal-2023-lambada, yao2023tree, wang2025stepwise}. Although these methods have achieved notable advancements, there are still limitations:

%%%%%%%%%%%%%%%%%%%%%%%%%%%%%%%%%%%%%%
\begin{figure*}[t]
\centering
\includegraphics[width=1\textwidth]{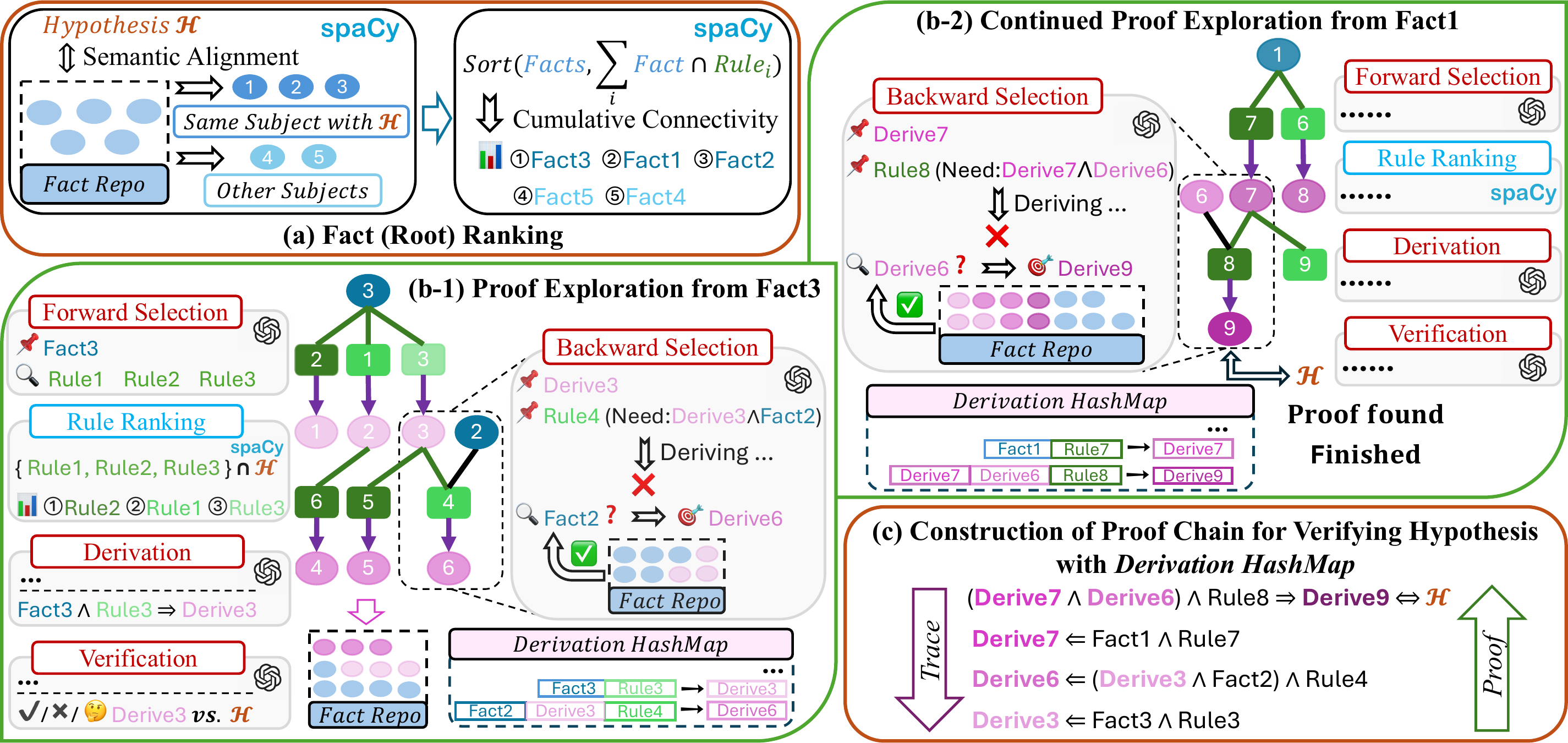}
\caption{The overview of LogicTree: (a) Fact (root) ranking; (b) Tree-by-tree search for proof exploration: (b-1) Proof exploration from the top-ranked fact (Fact3) which has the highest cumulative connectivity with rules, (b-2) Continued proof exploration from the next ranked fact (Fact1); (c) Construction of proof chain. The framework consists of (i) two caches: Fact Repository (Fact Repo) and
Derivation HashMap; (ii) four LLM-based modules: Forward Selection, Backward Selection, Derivation, Verification. Additionally, we leverage \texttt{spaCy} for fact and rule ranking. Within a tree: a blue oval represents a given fact; a green rectangle represents a rule; a purple oval represents a derived fact. LogicTree on an example from ProofWriter ~\citep{tafjord2020proofwriter} is shown in Figure~\ref{fig5}.}
\label{fig1}
\end{figure*}
%%%%%%%%%%%%%%%%%%%%%%%%%%%%%%%%%%%%%%

\noindent(1) Difficulties in maintaining logical coherence and in effectively utilizing derived knowledge obstruct progressive proof construction. In some iterative approaches ~\citep{creswell2023selectioninference, zhang2025cumulative}, reasoning steps are not required to build directly on prior derivations, which may disrupt logical coherence, hinder deep reasoning and cause redundancy ~\citep{wang2025stepwise}. While tree-based method ~\citep{yao2023tree} mitigates this issue, it lacks mechanisms to share derived knowledge across branches, potentially leading to reasoning stagnation ~\citep{sun-etal-2024-determlr}.

\noindent(2) Combinatorial complexity hinders precise premise selection which is essential for accurate stepwise reasoning. At each step, the model must identify the right combination of facts and rules from premise space for subsequent inference. The combinatorial search increases the risk of imprecise selection, which results in failed or inaccurate inference ~\citep{kazemi-etal-2023-lambada, liu2024concise}.

\noindent(3) Employing LLMs for proof planning ~\citep{wang2023plan} may be ineffective for complex logical reasoning, as such tasks require extensive and adaptive exploration. This often renders LLMs' planning unreliable and uninterpretable ~\citep{saparov2023language, kambhampati2024position}. Meanwhile, low-computation LLM-free heuristics may be sufficiently effective for strategic proof search, yet they remain largely overlooked.

To address these limitations, we propose LogicTree, a novel inference-time modular framework for structured proof exploration. The overview of our framework is shown in Figure~\ref{fig1}. LogicTree includes four LLM-based modules: \textit{Forward Selection}, \textit{Backward Selection}, \textit{Derivation} and \textit{Verification}, which are embedded in tree structure. Additionally, we incorporate Fact Repository and Derivation HashMap into LogicTree as cache components. Fact Repository is initialized with given facts and dynamically stores derived facts. It enables branches to access the given facts and the derivations from earlier branches, facilitating cross-branch information flow and effective utilization of historical knowledge throughout proof exploration.
%It is accessible across tree branches, enabling effective utilization of historical knowledge throughout proof exploration. 
Derivation HashMap records derived facts along with their derivation paths for a traceable reasoning process. We employ depth-first search (DFS) to orchestrate LLM modules and cache components, automating systematic proof search while ensuring logical coherence.

Furthermore, at each reasoning step, our framework decomposes the search for fact-rule combinations into \textit{Forward (rule) Selection} followed by \textit{Backward (fact) Selection}, reducing the complexity from combinatorial to linear. With this optimization, each selected rule-fact combination includes exactly one rule and its relevant fact(s), restricting inference to at most one derivation per step. This key improvement enhances reasoning granularity and enforces strict step-by-step reasoning, contributing to strengthened reasoning rigor.

% \textit{Forward Selection} identifies each relevant \textit{rule} by anchoring on a fact. A rule is selected if the fact fully or partially satisfies its conditions. When a rule's condition is partially satisfied, \textit{Backward Selection} searches the fact repository for missing \textit{fact} that completes the rule. Each rule-fact combination includes exactly one rule and its relevant fact(s), restricting each inference step to at most one derived fact.

Additionally, we introduce two heuristics leveraging \texttt{spaCy}\footnote{An open source NLP library (\url{https://spacy.io/}).} for premise prioritization: (1) \textit{Fact (root) ranking} for global ordering of tree search; (2) \textit{Rule ranking} at local level for early stopping in DFS. These LLM-free heuristics provide computationally efficient and interpretable strategies to accelerate proof-finding, avoiding blind and exhaustive search.

Our framework enables extensive exploration and fine reasoning granularity, optimally scaling reasoning length and inference-time computation ~\citep{snell2024scaling} to enhance logical reasoning capability. Our evaluation on five challenging logical reasoning benchmarks demonstrates that LogicTree significantly outperforms chain-of-thought (CoT) ~\cite{wei2022chain} and other modular methods in proof accuracy. Furthermore, within LogicTree framework, both Llama-3.3 70B ~\citep{dubey2024llama} and GPT-4o ~\citep{achiam2023gpt} surpass OpenAI’s o1-mini and o3-mini models. In-depth analysis reveals that our approach facilitates precise premise selection and accurate inference at each reasoning step while minimizing redundancy. 

% Additionally, our premise-prioritization strategies demonstrate improved efficiency in proof-finding by requiring fewer reasoning steps.

The main contributions of our work are:

\begin{itemize}[noitemsep,nolistsep]
\item We propose LogicTree, a novel inference-time framework that enables structured proof exploration while ensuring logical coherence. Additionally, we integrate cache components to effectively utilize historical knowledge and facilitate traceable reasoning process.

\vspace{3pt}
% \item We address the challenge of combinatorial complexity in premise search and enhance reasoning granularity, improving stepwise reasoning accuracy and reasoning rigor.
\item We address combinatorial complexity in premise search and enhance reasoning granularity, improving stepwise reasoning accuracy and strengthening overall reasoning rigor.
\vspace{3pt}
\item We introduce two LLM-free heuristics for premise prioritization, providing low-computation and interpretable strategies to improve proof-finding efficiency.
% in complex logical reasoning.

% providing low-computation and interpretable methods for strategic proof-finding in complex logical reasoning.
\end{itemize}

\section{Related Work}

\subsection{Reasoning through Strategic Prompting}

Pre-trained language models ~\citep{brown2020language, chowdhery2023palm, touvron2023llama} exhibit emergent reasoning abilities with increasing model scale. Strategic prompt engineering techniques, such as CoT ~\citep{wei2022chain, kojima2022large}, Auto-CoT ~\citep{zhang2023automatic}, self-consistency ~\citep{wang2023selfconsistency}, least-to-most ~\citep{zhou2023leasttomost}, help guide LLMs through intermediate reasoning steps, significantly improving LLM reasoning performance. However, the inherent simplicity of CoT and its variants, which is typically characterized by a left-to-right reasoning process with limited reasoning length, restricts their effectiveness in logical reasoning tasks that require exploration ~\citep{yao2023tree, xie2023self}.

\subsection{Inference-time Scaling for Reasoning}

Just as human may take more time to carefully analyze a complex question, enabling LLMs to refine their response with deliberate reasoning and increased inference-time computation is crucial for developing intelligent reasoning systems ~\citep{snell2024scaling, openai2024o1, chen2025towards}.

\noindent \textbf{Reasoning models trained via reinforcement learning (RL).}\quad Applying large-scale RL in LLM post-training phase has proven highly effective in enhancing reasoning abilities. It enables LLMs to develop reflection, self-correction, and long-chain reasoning skills for problem-solving ~\citep{openai2024o1, kumar2024training, shao2024deepseekmath, yeo2025demystifying}. Recently, DeepSeek-R1 ~\citep{guo2025deepseek} made significant breakthrough by achieving strong reasoning performance purely through RL, without the need for supervised fine-tuning (SFT).

\noindent \textbf{Modular inference without LLM parameter updates.}\quad Modular approach decomposes complex reasoning tasks into simpler sub-tasks, each assigned to specialized LLM modules implemented through few-shot prompting ~\citep{khot2023decomposed}. In logical reasoning, it involves two key modules that operate iteratively: premise selection and inference ~\citep{creswell2023selectioninference}. Extending from this foundation, Cumulative Reasoning ~\citep{zhang2025cumulative} integrates LLM verifier to validate reasoning steps. DetermLR ~\citep{sun-etal-2024-determlr} employs LLM scorer to prioritize relevant premises. SymbCoT ~\citep{xu2024faithful} and Aristotle ~\citep{xu2024aristotle} introduce LLM translator to convert natural language input into symbolic representations. Further research embeds LLM modules into topological structures, enabling deliberate problem solving ~\citep{yao2023tree, besta2024graph}.

% Current LLM-based methods for logical reasoning still struggle with structured exploration, maintaining logical coherence, and ensuring reasoning rigor in complex, multi-step tasks. 
Current LLM-based methods for logical reasoning still struggle to perform structured exploration while ensuring logical coherence and rigor in complex, multi-step reasoning tasks.
To address these challenges, we propose a novel inference-time modular approach that enables systematic and extensive proof exploration and enhances reasoning rigor.

\section{LogicTree for Logical Reasoning}
\subsection{Task Definition}

Logical reasoning aims to determine the truth value (\textit{true}, \textit{false}, or \textit{unknown}) of a \textit{hypothesis} $\mathcal{H}$ based on a set of premises consisting of \textit{facts} $\mathcal{F}$ and \textit{rules} $\mathcal{R}$ ~\citep{dowden2020logical}. An example is shown in Figure~\ref{fig5}.
% \vspace{3pt}
% {\centering
% \small % or \footnotesize
% \begin{tabular}{@{}l@{}}
%     \textit{fact}: \text{Kevin is hungry.} \\[0.3ex]
%     \textit{rule}: \text{If a person is hungry, the person is uncomfortable.} \\[0.3ex]
%     \textit{hypothesis}: \text{Kevin is uncomfortable.}
% \end{tabular}
% \par % Needed after \centering in a group to restore proper spacing
% }
% \vspace{3pt}
\noindent Formally,
$\mathcal{F} = \{ f_{i} \mid i = 1, 2, \dots, N_\mathcal{F} \}$, where each \( f_i \) represents a definitive statement within the reasoning system.
$\mathcal{R} = \{ r_{i} \mid i = 1, 2, \dots, N_\mathcal{R} \}$, where each \( r_i \) represents a conditional statement that defines a logical relationship between facts and inferred conclusions.
The reasoning process applies standard logical operators, including: Negation (\(\neg\)), Conjunction (\(\land\)), Disjunction (\(\lor\)), Implication (\(\Rightarrow\)), Equivalence (\(\Leftrightarrow\)). We define the set of intermediate derived facts as $\mathcal{D} = \{ d_{i} \mid i = 1, 2, \dots, N_\mathcal{D} \}$.

% Given the available facts, rules, and hypothesis as input, we propose LogicTree to structure the logical reasoning process. We will elaborate on the proposed framework in the following sections: components of LogicTree framework (\S~\ref{3.2}); LLM-free premise-prioritization heuristics (\S~\ref{3.3}); the LogicTree algorithm (\S~\ref{3.4}).

\subsection{Components of LogicTree Framework}
\label{3.2}

As shown in Figure~\ref{fig1}, the LogicTree framework includes (1) two caches: \textbf{Fact Repository} and \textbf{Derivation HashMap}; (2) four LLM-based modules: \textbf{Forward Selection}, \textbf{Derivation}, \textbf{Backward Selection}, and \textbf{Verification}, each implemented by few-shot prompting. The specific prompts for each module, along with example inputs and outputs, are provided in Appendix~\ref{example_and_prompt}.

\noindent\textbf{Fact Repository and Derivation HashMap.}\quad Fact Repository is initialized with given facts $\mathcal{F}$ and continuously stores derived facts $\mathcal{D}$. It enables tree branches to access the given facts and earlier derivations. This facilitates cross-branch information flow and effective utilization of historical knowledge throughout proof exploration.
Additionally, it checks whether a newly derived fact from a tree branch is unique among those already stored. If not, the branch is marked as a dead end to avoid redundancy and circular reasoning. Derivation HashMap stores derived facts as keys and their derivation paths as values, enabling a traceable reasoning process. Upon proof completion, the proof chain is reconstructed bottom-up, starting from the final path that verifies the hypothesis. If a fact in the path is found in the HashMap (i.e., it is derived rather than given), its associated derivation path is retrieved. This process occurs iteratively, constructing a streamlined proof as shown in Figure~\ref{fig1}c.

\noindent\textbf{Forward Selection Module.}\quad Based on a fact (either $f_{i}$ or $d_{i}$), this module selects all the relevant \textbf{rules} from the given rule set $\mathcal{R}$. A rule is considered relevant if its condition(s) are fully or partially satisfied by the fact. Each selected rule is added as a child node of the fact, forming parallel branches in the tree structure.

\noindent\textbf{Derivation Module.}\quad Along each branch, this module performs a strict one-step derivation using the current leaf rule and its parent fact. A successful derivation occurs if the fact fully satisfies the rule's condition. 
% The newly derived fact is added as a child node to the tree only if it is not already present in Fact Repository (to avoid redundancy). 
% If it is unique, it is then stored in Fact Repository, and its fact-derivation pair is cached in Derivation HashMap. 
If the derivation fails, it results from one of the two reasons: (1) the fact does not satisfy the rule’s condition at all, i.e., the rule was incorrectly selected by Forward Selection Module; (2) the fact partially satisfies the rule’s conditions, with some required fact(s) still missing. In the first case, the branch is marked as a dead end. In the second case, where conjunctive reasoning (e.g., $f_1 \land f_2 \land r_1 \Rightarrow\ d_1$) is required, the branch is marked as a \textit{pseudo} dead-end, where the missing fact(s) may still be retrievable. %we rely on Backward Selection Module to check if the missing condition (fact) can be retrieved to complete the rule and proceed inference.

\noindent\textbf{Backward Selection Module.}\quad If a branch is marked as a pseudo dead-end, this module is queried to attempt rule completion and resolve the stagnation. This module uses the current fact-rule pair as a pivot to identify the missing \textbf{fact(s)} required for derivation. It then searches Fact Repository to determine their availability. The missing fact(s) may be a given fact $f_{i}$ (Figure~\ref{fig1}b-1) or a derived fact $d_{i}$ from an earlier branch (Figure~\ref{fig1}b-2). If the missing fact(s) are available, the rule together with its supplemented relevant facts are then sent to Derivation Module to re-attempt derivation. If not, the branch is marked as dead end. 

% Note, backward selection differs from backtracking. Backtracking discards the current branch when a dead end is reached and explores alternative paths, while backward selection attempts to find the missing conditions (facts) to proceed inference in the current branch.

% This process consists of two steps: First the module uses the fact-rule pair in the current branch as a pivot to determine the missing condition required for inference. Second, it searches Fact Repository to check if the missing condition is available. The missing condition could be a given fact $f_{i}$ (Figure~\ref{fig1}b-1) or a previously derived fact $d_{i}$ from an earlier branch (Figure~\ref{fig1}b-2).

\noindent\textbf{Verification Module.}\quad After each successful derivation, this module evaluates the derived fact $d_{i}$ against the hypothesis $\mathcal{H}$ to determine if the proof is complete. If the derived fact is equivalent to or directly contradicts the hypothesis, the proof is concluded; otherwise, proof exploration continues.

\subsection{LLM-free Premise Prioritization}
\label{3.3}
% Prior studies propose employing LLMs for overall planning ~\citep{wang2023plan, xu2024faithful} and for prioritizing local reasoning steps ~\citep{yao2023tree, hao2023reasoning}. However, these LLM-based approaches may be less effective for complex logical reasoning. The nonlinear nature of such tasks, which may lead to exponential growth of local inference paths, makes it difficult to devise a precise overall plan. When prioritizing local reasoning steps, Tree-of-Thought (ToT) ~\citep{yao2023tree} lacks clear criterion in its prompts, leading to inconsistent and uninterpretable decisions. These limitations weaken the effectiveness of LLMs for planning and prioritization, resulting in inefficient use of computational resource. Meanwhile, simple LLM-free heuristics for premise prioritization have been largely overlooked. To address this gap, 

We introduce two heuristics leveraging \texttt{spaCy} for premise (fact and rule) prioritization, which provide low-computation and interpretable strategies to improve proof-finding efficiency.

\noindent\textbf{Fact (root) ranking for global ordering of tree search.}\quad In LogicTree framework, each given fact $f_i$ serves as the root of a tree, and trees are explored sequentially until the proof is found. 
%Fact ranking is performed prior to proof exploration. 
As shown in Figure~\ref{fig1}a, we first apply a semantic alignment step to prioritize facts that have the same \textit{subject} with the hypothesis $\mathcal{H}$ as tree roots. To further rank facts, we define \textbf{cumulative connectivity} between a fact $f_i$ and the rule set $\mathcal{R}$, which is the sum of \textbf{semantic overlap} between $f_i$ and each rule $r_i \in \mathcal{R}$. It approximates how many reasoning branches the root fact can initiate through its relevant rules. Facts with zero connectivity are discarded, as they cannot contribute to any derivation. Facts with higher connectivity are prioritized for opening more reasoning paths and higher likelihood of proof discovery in earlier-explored trees. We conduct subject alignment and compute semantic overlap using \texttt{spaCy}’s efficient dependency parsing.

\noindent\textbf{Rule ranking at local level for early stopping.} \quad 
% Rule ranking sorts the rules selected by Forward Selection Module based on each rule's \textbf{semantic overlap} with hypothesis $\mathcal{H}$. 
After each Forward Selection, the selected rules are ranked based on each rule's \textbf{semantic overlap} with hypothesis $\mathcal{H}$.
This prioritization directs Derivation Module to first apply the rule $r_i$ whose derivation is most likely to verify hypothesis $\mathcal{H}$, facilitating early stopping. For example, to verify $\mathcal{H}$: "\textit{Kevin is uncomfortable.}", a rule $r_i$ such as: "\textit{If ..., the person is uncomfortable.}" would be prioritized.

The computations for semantic overlap and cumulative connectivity are provided in Appendix~\ref{compute}.

\subsection{LogicTree Algorithm}
\label{3.4}
We employ iterative depth-first search (DFS) algorithm within LogicTree framework to automate systematic exploration as provided in Algorithm~\ref{algo1}. 

Initially, the algorithm uses \textit{Verification} to check if hypothesis $\mathcal{H}$ can be directly verified from given facts $\mathcal{F}$. If $\mathcal{H}$ is explicitly confirmed or refuted, the algorithm terminates and returns \textit{True} or \textit{False}, respectively. Otherwise, it proceeds with tree search.

As a preliminary, Fact Repository and Derivation HashMap are initialized, and the given facts are ranked using Fact Ranking heuristic. The algorithm starts with the top-ranked fact, which serves as the root of the first tree. Then, \textit{Forward Selection} is called to select relevant rules, which are subsequently ranked using Rule Ranking heuristic. Along each fact-rule branch, one-step inference is conducted. As shown in Algorithm~\ref{algo2}, the inference process encapsulates calls to \textit{Derivation} and, if necessary, \textit{Backward Selection}. \textit{Backward Selection} is triggered when the output of \textit{Derivation} indicates a pseudo dead-end. If \textit{Backward Selection} successfully retrieves the missing fact(s) from Fact Repository, then a secondary query to \textit{Derivation} is performed. 
Together, this modular process (i) decomposes the search for fact-rule combination, reducing complexity from combinatorial to linear (more analysis in Appendix~\ref{linear} and Table~\ref{table_complexity}); (ii) ensures each reasoning step involves exactly one rule and its relevant fact(s), producing at most one derived fact per step; (iii) avoids reasoning stagnation by attempting to resolve pseudo dead-ends.

After each inference, \textit{Verification} evaluates the result to determine whether it concludes the proof, enabling early stopping (Algorithm~\ref{algo3}). If the result indicates an underivable or redundant (i.e., already
in Fact Repository) outcome, DFS backtracks to explore the next branch. Otherwise, the derived fact $d_i$ is appended to the tree for further expansion.

The next iteration begins from derived fact $d_i$, with LLM modules reset before the next call. By building each step upon prior derivations, our framework maintains logical coherence. 
% Instead of restricting the tree’s depth or breadth, we set a step limit. 
Once a tree is fully explored, the algorithm proceeds to the next tree, using the next ranked fact as the root. 
To avoid excessively long reasoning, we set an LLM query limit on the reasoning process.
If all trees are explored or the query limit is reached (whichever occurs first) without verifying the hypothesis $\mathcal{H}$, the algorithm terminates and returns \textit{Unknown}.

% Overall, LogicTree ensures logical coherence by building each step upon prior derivations. The dynamic fact repository, combined with \textit{Backward Selection}, enhances the effective utilization of historical derivations during the tree search, preventing reasoning stagnation by resolving pseudo dead-ends. Additionally, it enables redundancy checks and helps avoid circular reasoning. \textit{Forward (rule) Selection} and \textit{Backward (fact) Selection} together decompose premise search, reducing its complexity from combinatorial to linear in order to identify the precise combination of facts and rules for inference. Further, the premise-prioritization heuristics offer computationally efficient and explainable strategies to accelerate proof-finding.

%%%%%%%%%%%%%%%%%%%%%%%%%%%%%%%%%%%%%%%%%%%%%%%%%
\begin{table*}
  \small
  \begin{center}
  \renewcommand{\arraystretch}{1} % Adjust row spacing
  \resizebox{1\textwidth}{!}{
\begin{tabular}{c c ccccc c}
    \toprule
    \multirow{2.5}{*}{\textbf{Model}} & \multirow{2.5}{*}{\textbf{Method}} & \multicolumn{5}{c}{\textbf{Dataset}} & \multirow{2.5}{*}{\textbf{Avg.\textsuperscript{\dag}}} \\
    \cmidrule(lr){3-7}
    & & \text{LogicNLI} & \text{ParaRules} & \text{PrOntoQA-OOD} & \text{ProofWriter} & \text{RobustLR} & \\
    \midrule

    %%%%%%%%
    \multirow{6}{*}{\centering \textbf{GPT-4o-mini}} 
      & CoT     & 38.0 & 48.3 & 55.0 & 51.8 & 62.1 & 51.2 \\
      
      \noalign{\vskip 1pt} % Adds small space before
      \cdashline{2-8}
      \noalign{\vskip 2pt} % Adds small space after

      & SI      & 46.0 & 51.3 & 72.5 & 55.3 & 60.4 & 55.8 \\
      & CR      & 42.7 & 54.0 & 75.0 & 49.7 & 70.0 & 56.1 \\
      & LAMBADA & \underline{54.7} & 62.0 & \underline{75.5} & 68.0 & 66.3 & 65.5 \\
      & ToT     & 51.3 & \underline{64.3} & 65.5 & \underline{70.3} & \underline{72.9} & \underline{66.5} \\
      
      \noalign{\vskip 1pt} % Adds small space before
      \cdashline{2-8}
      \noalign{\vskip 2pt} % Adds small space after
          
      \rowcolor{gray!20} % Light gray background
      & \textbf{LogicTree} & \textbf{58.0} & \textbf{68.7} & \textbf{87.5} & \textbf{78.8} & \textbf{87.1} & \textbf{75.7}\\
    %%%%%%%%

    \midrule

    %%%%%%%%
    \multirow{6}{*}{\centering \textbf{GPT-4o}} 
      & CoT     & 51.3 & 69.0 & 83.0 & 73.5 & 79.6 & 72.0 \\
      
      \noalign{\vskip 1pt} % Adds small space before
      \cdashline{2-8}
      \noalign{\vskip 2pt} % Adds small space after

      & SI      & 48.0 & 71.0 & 91.5 & 68.0 & 71.3 & 70.4 \\
      & CR      & 54.0 & \underline{75.3} & 91.5 & 75.3 & 76.7 & 75.5 \\
      & LAMBADA & 68.0 & 73.3 & \underline{93.5} & 86.7 & 88.3 & 81.6 \\
      & ToT     & \underline{69.3} & 75.0 & 86.5 & \underline{91.0} & \underline{89.2} & \underline{83.1} \\
      
      \noalign{\vskip 1pt} % Adds small space before
      \cdashline{2-8}
      \noalign{\vskip 2pt} % Adds small space after
          
      \rowcolor{gray!20} % Light gray background
      & \textbf{LogicTree} & \textbf{78.7} & \textbf{96.3} & \textbf{99.0} & \textbf{97.0} & \textbf{97.9} & \textbf{95.6}\\
    %%%%%%%%

    \midrule

    %%%%%%%%
    \multirow{6}{*}{\centering \textbf{ Llama-3.3 70B}} 
      & CoT     & 46.7 & 70.8 & 88.5 & 75.5 & 80.0 & 73.6 \\
      
      \noalign{\vskip 1pt} % Adds small space before
      \cdashline{2-8}
      \noalign{\vskip 2pt} % Adds small space after

      & SI      & 52.0 & 74.7 & 92.5 & 61.7 & 76.3 & 70.6 \\
      & CR      & 53.3 & 76.7 & \underline{93.0} & 73.3 & 70.8 & 74.6 \\
      & LAMBADA & 66.7 & 78.3 & 91.0 & 81.7 & \underline{87.1} & 81.1 \\
      & ToT     & \underline{69.0} & \underline{79.7} & 90.5 & \underline{87.7} & 85.4 & \underline{83.4} \\
      
      \noalign{\vskip 1pt} % Adds small space before
      \cdashline{2-8}
      \noalign{\vskip 2pt} % Adds small space after
          
      \rowcolor{gray!20} % Light gray background
      & \textbf{LogicTree} & \textbf{74.7} & \textbf{92.3} & \textbf{97.0} & \textbf{95.8} & \textbf{97.5} & \textbf{93.2}\\
    %%%%%%%%

    \bottomrule
    \end{tabular}
  }
  \end{center}
  \caption{Proof accuracy of different methods across five logical reasoning datasets on GPT-4o-mini, GPT-4o, and Llama-3.3 70B. The highest accuracy in each case is in bold; the second-highest is underlined. Avg.\textsuperscript{\dag} is calculated as the number of correctly proved examples divided by the total number of examples across all five datasets.}
  \label{table1}
\end{table*}
%%%%%%%%%%%%%%%%%%%%%%%%%%%%%%%%%%%%%%%%%%%%%%%%%

%%%%%%%%%%%%%%%%%%%%%%%%%%%%%%%%%%%%%%
\begin{figure*}[t]
\centering
\includegraphics[width=1\textwidth]{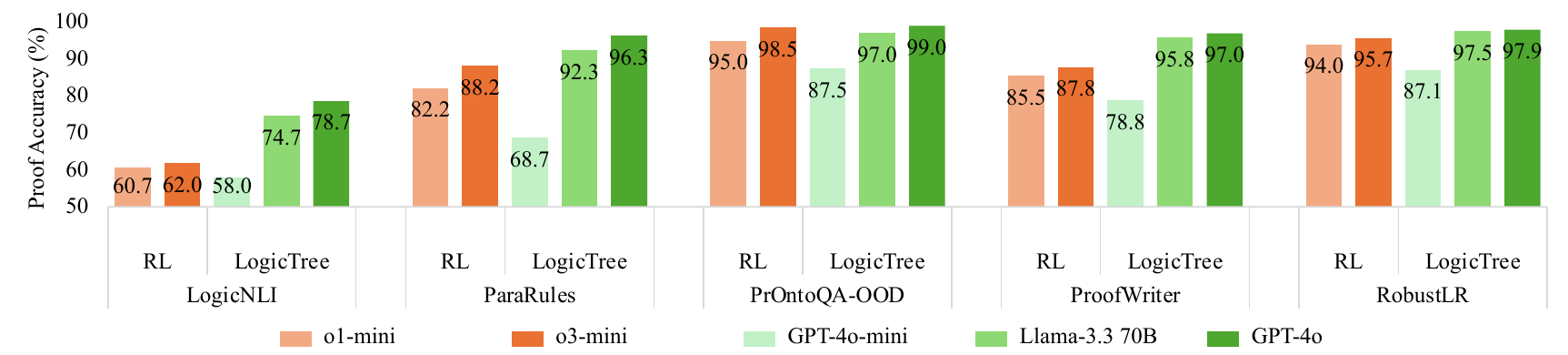}
\caption{Performance comparison between general LLMs (GPT-4o-mini, GPT-4o, Llama-3.3 70B) applied within LogicTree and RL-trained reasoning models (o1-mini, o3-mini).}
\label{fig2}
\end{figure*}
%%%%%%%%%%%%%%%%%%%%%%%%%%%%%%%%%%%%%%

\section{Experiments}
\subsection{Experimental Setup}

\textbf{Datasets.}\quad We evaluate our framework on five multi-step logical reasoning datasets: 
\textbf{RobustLR} ~\citep{sanyal-etal-2022-robustlr}, 
\textbf{PrOntoQA-OOD} ~\citep{saparov2023testing}, 
\textbf{ProofWriter} ~\citep{tafjord2020proofwriter}, 
\textbf{ParaRules} ~\citep{ijcai2020p0537}, 
\textbf{LogicNLI} ~\citep{tian-etal-2021-diagnosing}.
For all examples in our experiments, hypothesis concludes as \textit{True}, \textit{False}, or \textit{Unknown}. More details on logical reasoning datasets are provided in Appendix~\ref{dataset}. 
To further assess the broader adaptability of our framework, we extend it to mathematical reasoning datasets, as demonstrated in Appendix~\ref{math_extension}.
% (Appendix~\ref{math_extension} shows the extension of our framework to mathematical reasoning datasets.)
% We also apply our framework to mathematical reasoning task in Appendix~\ref{math_extension}.

\noindent\textbf{Baselines.}\quad To compare our framework with existing LLM-based reasoning methods, we select baselines from three categories:

\begin{itemize}[leftmargin=10pt,noitemsep,nolistsep]
    \vspace{2pt}
    \item \textit{Strategic LLM prompting}: \textbf{CoT}~\citep{wei2022chain} prompts the model to generate intermediate reasoning steps before providing final answers.
    \vspace{2pt}
    \item \textit{Modular approaches}: \textbf{SI} (Selection-Inference) ~\citep{creswell2023selectioninference} adpots selection and inference modules for iterative reasoning. \textbf{CR} (Cumulative Reasoning) ~\citep{zhang2025cumulative} introduces a cumulative process of generating new propositions to reach the answer. \textbf{ToT} (Tree-of-Thought) ~\citep{yao2023tree} leverages tree-search algorithm for deliberate reasoning. \textbf{LAMBADA} ~\citep{kazemi-etal-2023-lambada} develops a backward chaining approach for automated reasoning.
    \vspace{2pt}
    \item \textit{RL-trained reasoning models}: \textbf{o1-mini} ~\citep{openai2024o1} and \textbf{o3-mini} \citep{openai2025o3mini}  model.
    \vspace{2pt}
\end{itemize}

\noindent\textbf{Models.}\quad 
% Our proposed framework places no restrictions on the choice of LLMs. Here, we separately employ GPT-4o-mini, GPT-4o, ~\citep{achiam2023gpt} and Llama-3.3 70B ~\citep{dubey2024llama} within our framework. For a fair comparison, we reproduce CoT and other modular approaches using the same models. Further details on models are provided in Appendix~\ref{model}.
Our framework places no restrictions on the choice of LLMs. Here, we separately employ GPT-4o-mini, GPT-4o~\citep{achiam2023gpt}, and Llama-3.3 70B ~\citep{dubey2024llama} within our framework. We reproduce CoT and other modular approaches using the same models for comparison. Further details on models are in Appendix~\ref{model}.

\noindent\textbf{Evaluate reasoning accuracy.}\quad In logical reasoning, correct label prediction (\textit{True}, \textit{False}, or \textit{Unknown}) does not necessarily indicate correct reasoning, as models may arrive at the correct conclusion through hallucinated premises or spurious correlations ~\citep{kazemi-etal-2023-lambada, liu2023transformers}. Similar to ~\citet{saparov2023language}, we use \textbf{proof accuracy} for rigorous evaluation. We manually verify each example by focusing on the reasoning chain that verifies the hypothesis within the entire reasoning trace. A proof is considered correct if every step in this chain is valid, while the validity of other reasoning paths is disregarded.

%%%%%%%%%%%%%%%%%%%%%%%%%%%%%%%%%%%%%%
\begin{figure*}[t]
\centering
\includegraphics[width=1\textwidth]{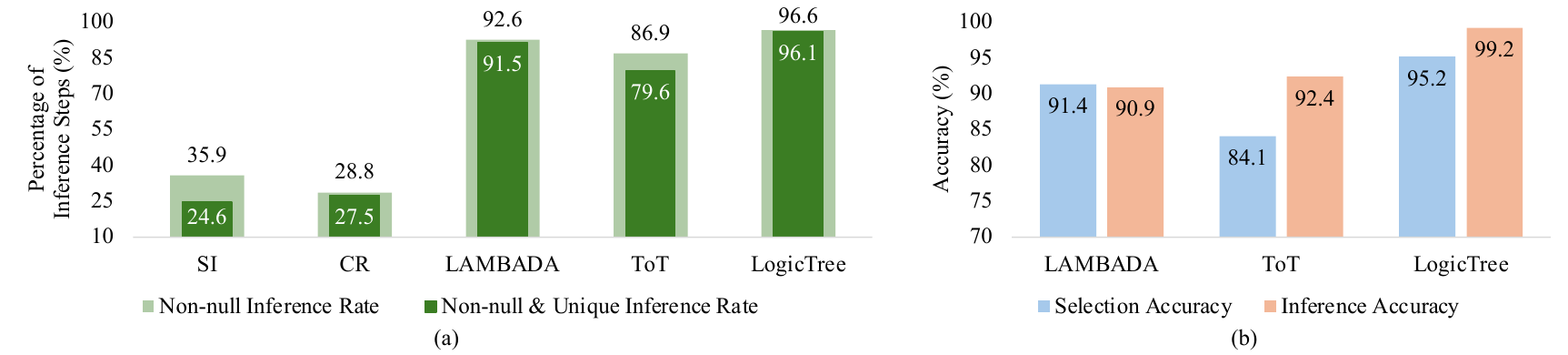}
\caption{Step-level metrics (\S~\ref{5.1}) across different methods. (a) Non-null inference rate (the outer bars) and non-null \& unique inference rate (the inner bars). (b) Selection accuracy and inference accuracy, evaluated only on tree-based methods. All metrics are manually evaluated on GPT-4o’s outputs for 100 examples from ProofWriter.}
\label{fig3}
\end{figure*}
%%%%%%%%%%%%%%%%%%%%%%%%%%%%%%%%%%%%%%

\subsection{Main Results}
As shown in Table~\ref{table1}, our proposed LogicTree consistently outperforms CoT and other modular approaches across all five datasets. Specifically, our method surpasses CoT significantly, with average performance gains of 24.5\%, 23.6\%, and 19.6\% on GPT-4o-mini, GPT-4o, and Llama-3.3 70B, respectively. Compared to ToT, the strongest among other modular methods, our framework achieves average improvements of 9.2\%, 12.5\%, and 9.8\% on the same models. 
On ParaRules, PrOntoQA-OOD, ProofWriter, and RobustLR datasets, our framework achieves near-perfect proof accuracy with GPT-4o, highlighting its strength in logical reasoning.
This strength generalizes across different levels of task difficulty, as shown in Figure~\ref{fig7}. 
% Furthermore, LogicTree demonstrates superior performance over RL-trained reasoning models, o1-mini and o3-mini, as presented in Figure~\ref{fig2}. With Llama-3.3 70B, our method yields 8.7\% and 5.2\% higher accuracy on average; with GPT-4o, it achieves average gains of 11.1\% and 7.6\%, respectively.
Furthermore, when applied within LogicTree, Llama-3.3 70B and GPT-4o outperform RL-trained reasoning models, o1-mini and o3-mini, as shown in Figure~\ref{fig2}.
With Llama-3.3 70B, our method yields 8.7\% and 5.2\% higher accuracy on average; with GPT-4o, the average gains are 11.1\% and 7.6\%, respectively.

\section{Further Analysis}

Figure~\ref{fig6} schematically illustrates how baseline approaches perform logical reasoning with LLMs, which facilitates in-depth performance analysis.

\subsection{Factors Impacting Proof Accuracy}
\label{5.1}
To explain the effectiveness of our framework, we define the following step-level metrics:

\noindent 1. \textit{Non-null Inference Rate}: The percentage of inference steps that result in derived facts.

\noindent 2. \textit{Non-null \& Unique Inference Rate}: The percentage of inference steps that generate new facts (i.e., not previously derived).

\noindent 3. \textit{Selection Accuracy}: The percentage of selection steps where the selected premises are logically relevant to the parent node during tree expansion.

\noindent 4. \textit{Inference Accuracy}: The percentage of inference steps that are logically correct given the selected premises.

\noindent\textbf{Logical coherence.}\quad Tree-based frameworks (ToT, LAMBADA, LogicTree) exhibit significantly higher performance than SI and CR due to better maintenance of logical coherence.
As shown in Figure~\ref{fig6}, SI and CR begin each iteration from the updated premise set rather than building directly on prior derivations, disrupting logical coherence. This disruption breaks the continuity of reasoning, resulting in the loss of the logical "pivot" (i.e., prior derivation) needed to guide premise selection. For SI, without this anchor, identifying logically relevant fact-rule combinations becomes difficult, resulting in frequent failed (null) inferences. Additionally, the lack of coherence limits awareness of previous derivations, leading to repeated re-derivation and redundancy. These issues are reflected by SI's low \textit{non-null \& unique inference rate} in Figure~\ref{fig3}a.
CR adopts random combination for premise selection, resulting in an even lower \textit{non-null inference rate} (Figure~\ref{fig3}a) due to irrelevant selected premises. Under a fixed iteration budget, failed and redundant steps stall logical progression and ultimately render the proof incomplete.

\noindent\textbf{Premise selection accuracy in tree search.}\quad Although ToT builds each reasoning step on prior derivations, it still faces combinatorial search complexity. In conjunctive reasoning scenario (e.g., $f_1 \land f_2 \land r_1 \Rightarrow\ d_1$), it requires searching for relevant fact-rule combination ($f_2 \land r_1$) for a parent fact node ($f_1$), making precise selection challenging. Also, to accommodate such search process, ToT does not constrain the number of selected premises per branch (Figure~\ref{fig6}d), increasing the risk of selecting irrelevant premises (i.e., distractions). Together, these factors reduce \textit{selection accuracy} and subsequently lead to failed or inaccurate inferences.

\noindent\textbf{Forward vs. Backward tree search strategies.}\quad LAMBADA (backward reasoning) starts from the hypothesis and checks each rule to determine its applicability. This method inherently avoids combinatorial search (Table~\ref{table_complexity}), leading to higher \textit{selection accuracy}. However, despite the challenge of combinatorial search, forward reasoning (ToT) achieves higher \textit{inference accuracy} than backward reasoning (Figure~\ref{fig3}b). This may be attributed to the prevalence of forward logical flow in pre-training corpus and the autoregressive nature of LLMs, which favors reasoning from premises to conclusions.

Our framework adopts forward reasoning to leverage its aforementioned advantage, while decomposing premise selection to address its search complexity (Table~\ref{table_complexity}), effectively improving both \textit{selection accuracy} and \textit{inference accuracy}, as shown in Figure~\ref{fig3}b. Another key reason our framework outperforms ToT is that ToT lacks a mechanism to leverage derived facts from earlier branches, which may lead to reasoning stagnation (the scenario in Figure~\ref{fig1}b-2). Our framework addresses this issue through incorporating Fact Repository (\S~\ref{3.2}). The impact of this component is evaluated in Table~\ref{table_fact_repo}.

We further conduct error analysis on CoT, o1-mini, o3-mini, and our framework in Appendix~\ref{error}.

%%%%%%%%%%%%%%%%%%%%%%%%%%%%%%%%%%%%%%
\begin{figure}[ht]
\centering
\includegraphics[width=1\linewidth]{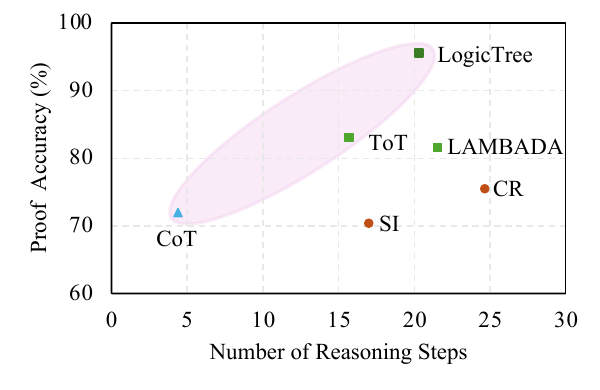}
\caption{Proof accuracy vs. reasoning steps, averaged across five datasets for GPT-4o. The shaded area illustrates that our framework optimally scales inference-time computation to achieve higher proof accuracy.}
\label{fig4}
\end{figure}
%%%%%%%%%%%%%%%%%%%%%%%%%%%%%%%%%%%%%%

\subsection{Scaling of Reasoning Length}

\textbf{Proof accuracy vs. reasoning length.}\quad To assess the impact of long reasoning on solving complex logical tasks and compare its effectiveness across different approaches, we measure the average number of reasoning steps for each approach across five datasets, with details in Appendix~\ref{steps}. The corresponding proof accuracy and average reasoning steps for each approach are presented in Figure~\ref{fig4}. Insufficient reasoning of CoT in complex tasks limits its performance. SI and CR, as analyzed in \S~\ref{5.1}, suffer from high proportion of redundant and failed inferences, which undermine the effectiveness of long reasoning. LAMBADA (backward reasoning) demonstrates more reasoning steps and lower proof accuracy compared to forward reasoning (ToT and LogicTree). Additional analysis comparing forward and backward reasoning is provided in Appendix~\ref{for_vs_back} Table~\ref{table_forward_backward}. Compared to ToT, our framework requires more reasoning steps for two main reasons: (1) Our framework decomposes the combinatorial premise search, leading to more steps; (2) In ToT, multiple derivations can occur within a single step (Figure~\ref{fig6}d), whereas our framework restricts each step to at most one derivation. The enhanced reasoning granularity ensures strict step-by-step reasoning, optimally increasing reasoning length and leveraging additional inference-time computation to achieve higher proof accuracy, as shown in the shaded areas of Figure~\ref{fig4} and Figure~\ref{fig8}.

We additionally report the average token cost and inference time for all methods in Table~\ref{output_token_time}, complementing the reasoning-step counts. Figure~\ref{token_budget} further illustrates the trade-off between accuracy and inference cost by showing proof accuracy under varying token budgets. The results indicate that (1) LogicTree consistently outperforms the baseline approaches at every budget level, and (2) LogicTree achieves larger performance gains as the token budget increases, demonstrating more effective scaling with additional inference computation.

\begin{table}
    \small
    \begin{center}
    \renewcommand{\arraystretch}{1.1} % Adjust row spacing
    \resizebox{1\linewidth}{!}{
    \begin{tabular}{l ccc}
        \toprule
        & ParaRules & ProofWriter & RobustLR \\
        \midrule
        \multirow{2}{*}{\centering w/o prioritize} & 17.8 & 47.0 & 19.7 \\
        & {\color{gray} (93.0)} & {\color{gray} (91.7)} & {\color{gray} (94.6)} \\[2pt]
        
        \multirow{2}{*}{\centering LLM-based prioritize} & 11.6 & 30.6 & 17.5 \\
        & {\color{gray} (95.5)} & {\color{gray} (94.8)} & {\color{gray} (96.7)} \\[2pt]
        
        \multirow{2}{*}{\textbf{Proposed prioritize}} & \textbf{9.5} & \textbf{23.5} & \textbf{13.8} \\
        & \textbf{\color{gray} (96.3)} & \textbf{\color{gray} (97.0)} & \textbf{\color{gray} (97.9)} \\
        \bottomrule
    \end{tabular}}
    \end{center}
    \caption{Ablation results on GPT-4o: average reasoning steps and proof accuracy (gray, in parentheses).}
    \label{table2}
\end{table}

\noindent\textbf{Premise-prioritization heuristics for efficient scaling.}\quad We introduce two premise-prioritization heuristics for strategic proof exploration (\S~\ref{3.3}). To evaluate their impact on proof search efficiency, we conduct an ablation study across three scenarios: (1) without premise prioritization, where both facts and selected rules are sampled in a random order for exploration; (2) using LLM-based premise prioritization, where two LLM modules are applied: one for fact ranking and one for rule ranking, with details provided in Appendix~\ref{llm_for_prior}; and (3) using our proposed LLM-free heuristics. As shown in Table~\ref{table2}, our proposed heuristics facilitate fewer reasoning steps in proof-finding while attaining higher proof accuracy by avoiding the increased error risk associated with longer reasoning paths.

\section{Conclusion}

In this work, we propose LogicTree, a novel inference-time modular framework for logical reasoning. Our framework employs algorithm-guided search (DFS) to automate structured exploration while ensuring logical coherence. It incorporates caching mechanism to effectively utilize historical knowledge, preventing reasoning stagnation and minimizing redundancy. Furthermore, we address the combinatorial complexity of premise search and enhance reasoning granularity by restricting inference to at most one derivation per step. This improves stepwise reasoning accuracy and strengthens reasoning rigor. Additionally, we introduce LLM-free heuristics that provide low-computation, explainable strategies to improve proof search efficiency. Experimental results show that LogicTree optimally leverages inference-time scaling to achieve higher proof accuracy, surpassing other modular frameworks and reasoning models, highlighting its strength in logical reasoning.

\section*{Limitations}

While our framework demonstrates strong performance in logical reasoning tasks, it has some limitations that could open avenues for future work. 

First, we primarily evaluate our framework in the domain of logical reasoning, as it represents a distinct type of challenge in reasoning tasks that requires structured and extensive exploration. Our goal is to address this type of reasoning challenge, which often demands more deliberate reasoning. In future work, we plan to extend the framework to more complex domains such as theorem proving.

Second, our framework assumes that all premises (i.e., facts and rules) are explicitly provided. Future work could incorporate premise augmentation with plausible knowledge retrieved from LLM, rather than relying solely on the given premises. Additionally, when facts and rules are not clearly separated, an extra pre-processing step with assistance from LLM may be required~\citep{sun-etal-2024-determlr}. Also, our premise prioritization strategies rely on simple heuristics. Developing more advanced approaches for proof planning and premise prioritization remains an important direction for future research.

\section*{Ethics Statement and Broader Impact}
Our work adheres to the Code of Ethics. All utilized methods, models, and datasets are properly cited. The datasets used in our experiments are publicly available, and our research does not involve any private or sensitive information. We confirm that our use of datasets and LLMs aligns with their intended purposes and usage guidelines. 
A potential risk of our framework lies in the misuse of its outputs in high-stakes domains without sufficient validation or expert review, as LLMs cannot always guarantee fully correct outputs. Nevertheless, when properly applied, our framework contributes to the development of interpretable and automated reasoning systems. Our work has the potential to extend to real-world applications that require rigorous, multi-step decision-making.

\section*{Acknowledgments}
This work was supported in part by the Center for Co-Design of Cognitive Systems (CoCoSys), a Semiconductor Research Corporation (SRC) and DARPA-sponsored JUMP 2.0 center.

\bibliography{custom}

\begin{thebibliography}{48}
\providecommand{\natexlab}[1]{#1}

\bibitem[{Achiam et~al.(2023)Achiam, Adler, Agarwal, Ahmad, Akkaya, Aleman, Almeida, Altenschmidt, Altman, Anadkat et~al.}]{achiam2023gpt}
Josh Achiam, Steven Adler, Sandhini Agarwal, Lama Ahmad, Ilge Akkaya, Florencia~Leoni Aleman, Diogo Almeida, Janko Altenschmidt, Sam Altman, Shyamal Anadkat, and 1 others. 2023.
\newblock Gpt-4 technical report.
\newblock \emph{arXiv preprint arXiv:2303.08774}.

\bibitem[{Amini et~al.(2019)Amini, Gabriel, Lin, Koncel-Kedziorski, Choi, and Hajishirzi}]{amini-etal-2019-mathqa}
Aida Amini, Saadia Gabriel, Shanchuan Lin, Rik Koncel-Kedziorski, Yejin Choi, and Hannaneh Hajishirzi. 2019.
\newblock \href {https://doi.org/10.18653/v1/N19-1245} {{M}ath{QA}: Towards interpretable math word problem solving with operation-based formalisms}.
\newblock In \emph{Proceedings of the 2019 Conference of the North {A}merican Chapter of the Association for Computational Linguistics: Human Language Technologies, Volume 1 (Long and Short Papers)}, pages 2357--2367, Minneapolis, Minnesota. Association for Computational Linguistics.

\bibitem[{Besta et~al.(2024)Besta, Blach, Kubicek, Gerstenberger, Podstawski, Gianinazzi, Gajda, Lehmann, Niewiadomski, Nyczyk et~al.}]{besta2024graph}
Maciej Besta, Nils Blach, Ales Kubicek, Robert Gerstenberger, Michal Podstawski, Lukas Gianinazzi, Joanna Gajda, Tomasz Lehmann, Hubert Niewiadomski, Piotr Nyczyk, and 1 others. 2024.
\newblock Graph of thoughts: Solving elaborate problems with large language models.
\newblock In \emph{Proceedings of the AAAI Conference on Artificial Intelligence}, volume~38, pages 17682--17690.

\bibitem[{Brown et~al.(2020)Brown, Mann, Ryder, Subbiah, Kaplan, Dhariwal, Neelakantan, Shyam, Sastry, Askell et~al.}]{brown2020language}
Tom Brown, Benjamin Mann, Nick Ryder, Melanie Subbiah, Jared~D Kaplan, Prafulla Dhariwal, Arvind Neelakantan, Pranav Shyam, Girish Sastry, Amanda Askell, and 1 others. 2020.
\newblock Language models are few-shot learners.
\newblock \emph{Advances in neural information processing systems}, 33:1877--1901.

\bibitem[{Chen et~al.(2025)Chen, Qin, Liu, Peng, Guan, Wang, Hu, Zhou, Gao, and Che}]{chen2025towards}
Qiguang Chen, Libo Qin, Jinhao Liu, Dengyun Peng, Jiannan Guan, Peng Wang, Mengkang Hu, Yuhang Zhou, Te~Gao, and Wangxiang Che. 2025.
\newblock Towards reasoning era: A survey of long chain-of-thought for reasoning large language models.
\newblock \emph{arXiv preprint arXiv:2503.09567}.

\bibitem[{Chen et~al.(2024)Chen, Chi, Wang, and Zhou}]{pmlr-v235-chen24i}
Xinyun Chen, Ryan~Andrew Chi, Xuezhi Wang, and Denny Zhou. 2024.
\newblock Premise order matters in reasoning with large language models.
\newblock In \emph{Proceedings of the 41st International Conference on Machine Learning}, volume 235 of \emph{Proceedings of Machine Learning Research}, pages 6596--6620. PMLR.

\bibitem[{Chowdhery et~al.(2023)Chowdhery, Narang, Devlin, Bosma, Mishra, Roberts, Barham, Chung, Sutton, Gehrmann et~al.}]{chowdhery2023palm}
Aakanksha Chowdhery, Sharan Narang, Jacob Devlin, Maarten Bosma, Gaurav Mishra, Adam Roberts, Paul Barham, Hyung~Won Chung, Charles Sutton, Sebastian Gehrmann, and 1 others. 2023.
\newblock Palm: Scaling language modeling with pathways.
\newblock \emph{Journal of Machine Learning Research}, 24(240):1--113.

\bibitem[{Clark et~al.(2020)Clark, Tafjord, and Richardson}]{ijcai2020p0537}
Peter Clark, Oyvind Tafjord, and Kyle Richardson. 2020.
\newblock \href {https://doi.org/10.24963/ijcai.2020/537} {Transformers as soft reasoners over language}.
\newblock In \emph{Proceedings of the Twenty-Ninth International Joint Conference on Artificial Intelligence, {IJCAI-20}}, pages 3882--3890. International Joint Conferences on Artificial Intelligence Organization.
\newblock Main track.

\bibitem[{Cobbe et~al.(2021)Cobbe, Kosaraju, Bavarian, Chen, Jun, Kaiser, Plappert, Tworek, Hilton, Nakano et~al.}]{cobbe2021training}
Karl Cobbe, Vineet Kosaraju, Mohammad Bavarian, Mark Chen, Heewoo Jun, Lukasz Kaiser, Matthias Plappert, Jerry Tworek, Jacob Hilton, Reiichiro Nakano, and 1 others. 2021.
\newblock Training verifiers to solve math word problems.
\newblock \emph{arXiv preprint arXiv:2110.14168}.

\bibitem[{Creswell et~al.(2023)Creswell, Shanahan, and Higgins}]{creswell2023selectioninference}
Antonia Creswell, Murray Shanahan, and Irina Higgins. 2023.
\newblock \href {https://openreview.net/forum?id=3Pf3Wg6o-A4} {Selection-inference: Exploiting large language models for interpretable logical reasoning}.
\newblock In \emph{The Eleventh International Conference on Learning Representations}.

\bibitem[{Dowden(2020)}]{dowden2020logical}
Bradley~Harris Dowden. 2020.
\newblock \emph{Logical reasoning}.
\newblock Bradley Dowden.

\bibitem[{Dubey et~al.(2024)Dubey, Jauhri, Pandey, Kadian, Al-Dahle, Letman, Mathur, Schelten, Yang, Fan et~al.}]{dubey2024llama}
Abhimanyu Dubey, Abhinav Jauhri, Abhinav Pandey, Abhishek Kadian, Ahmad Al-Dahle, Aiesha Letman, Akhil Mathur, Alan Schelten, Amy Yang, Angela Fan, and 1 others. 2024.
\newblock The llama 3 herd of models.
\newblock \emph{arXiv preprint arXiv:2407.21783}.

\bibitem[{Guo et~al.(2025)Guo, Yang, Zhang, Song, Zhang, Xu, Zhu, Ma, Wang, Bi et~al.}]{guo2025deepseek}
Daya Guo, Dejian Yang, Haowei Zhang, Junxiao Song, Ruoyu Zhang, Runxin Xu, Qihao Zhu, Shirong Ma, Peiyi Wang, Xiao Bi, and 1 others. 2025.
\newblock Deepseek-r1: Incentivizing reasoning capability in llms via reinforcement learning.
\newblock \emph{arXiv preprint arXiv:2501.12948}.

\bibitem[{Kambhampati et~al.(2024)Kambhampati, Valmeekam, Guan, Verma, Stechly, Bhambri, Saldyt, and Murthy}]{kambhampati2024position}
Subbarao Kambhampati, Karthik Valmeekam, Lin Guan, Mudit Verma, Kaya Stechly, Siddhant Bhambri, Lucas~Paul Saldyt, and Anil~B Murthy. 2024.
\newblock Position: Llms can’t plan, but can help planning in llm-modulo frameworks.
\newblock In \emph{Forty-first International Conference on Machine Learning}.

\bibitem[{Kazemi et~al.(2023)Kazemi, Kim, Bhatia, Xu, and Ramachandran}]{kazemi-etal-2023-lambada}
Mehran Kazemi, Najoung Kim, Deepti Bhatia, Xin Xu, and Deepak Ramachandran. 2023.
\newblock \href {https://doi.org/10.18653/v1/2023.acl-long.361} {{LAMBADA}: Backward chaining for automated reasoning in natural language}.
\newblock In \emph{Proceedings of the 61st Annual Meeting of the Association for Computational Linguistics (Volume 1: Long Papers)}, Toronto, Canada. Association for Computational Linguistics.

\bibitem[{Khot et~al.(2023)Khot, Trivedi, Finlayson, Fu, Richardson, Clark, and Sabharwal}]{khot2023decomposed}
Tushar Khot, Harsh Trivedi, Matthew Finlayson, Yao Fu, Kyle Richardson, Peter Clark, and Ashish Sabharwal. 2023.
\newblock \href {https://openreview.net/forum?id=_nGgzQjzaRy} {Decomposed prompting: A modular approach for solving complex tasks}.
\newblock In \emph{The Eleventh International Conference on Learning Representations}.

\bibitem[{Kojima et~al.(2022)Kojima, Gu, Reid, Matsuo, and Iwasawa}]{kojima2022large}
Takeshi Kojima, Shixiang~Shane Gu, Machel Reid, Yutaka Matsuo, and Yusuke Iwasawa. 2022.
\newblock Large language models are zero-shot reasoners.
\newblock \emph{Advances in neural information processing systems}, 35:22199--22213.

\bibitem[{Kumar et~al.(2024)Kumar, Zhuang, Agarwal, Su, Co-Reyes, Singh, Baumli, Iqbal, Bishop, Roelofs et~al.}]{kumar2024training}
Aviral Kumar, Vincent Zhuang, Rishabh Agarwal, Yi~Su, John~D Co-Reyes, Avi Singh, Kate Baumli, Shariq Iqbal, Colton Bishop, Rebecca Roelofs, and 1 others. 2024.
\newblock Training language models to self-correct via reinforcement learning.
\newblock \emph{arXiv preprint arXiv:2409.12917}.

\bibitem[{Liu et~al.(2023)Liu, Ash, Goel, Krishnamurthy, and Zhang}]{liu2023transformers}
Bingbin Liu, Jordan~T. Ash, Surbhi Goel, Akshay Krishnamurthy, and Cyril Zhang. 2023.
\newblock \href {https://openreview.net/forum?id=De4FYqjFueZ} {Transformers learn shortcuts to automata}.
\newblock In \emph{The Eleventh International Conference on Learning Representations}.

\bibitem[{Liu et~al.(2025)Liu, Fu, Ding, Ning, Zhang, Liu, and Zhang}]{liu2025logical}
Hanmeng Liu, Zhizhang Fu, Mengru Ding, Ruoxi Ning, Chaoli Zhang, Xiaozhang Liu, and Yue Zhang. 2025.
\newblock Logical reasoning in large language models: A survey.
\newblock \emph{arXiv preprint arXiv:2502.09100}.

\bibitem[{Liu et~al.(2024)Liu, Yan, Shen, Xie, Wang, and Ye}]{liu2024concise}
Junjie Liu, Shaotian Yan, Chen Shen, Liang Xie, Wenxiao Wang, and Jieping Ye. 2024.
\newblock Concise and organized perception facilitates reasoning in large language models.
\newblock \emph{arXiv preprint arXiv:2310.03309}.

\bibitem[{OpenAI(2024{\natexlab{a}})}]{openai2024o1}
OpenAI. 2024{\natexlab{a}}.
\newblock \href {https://openai.com/index/learning-to-reason-with-llms/} {Learning to reason with llms}.

\bibitem[{OpenAI(2024{\natexlab{b}})}]{openai2024prompt}
OpenAI. 2024{\natexlab{b}}.
\newblock \href {https://platform.openai.com/docs/guides/reasoning-best-practices} {Reasoning best practices}.

\bibitem[{OpenAI(2025)}]{openai2025o3mini}
OpenAI. 2025.
\newblock \href {https://openai.com/index/openai-o3-mini/} {Openai o3-mini}.

\bibitem[{Sanyal et~al.(2022)Sanyal, Liao, and Ren}]{sanyal-etal-2022-robustlr}
Soumya Sanyal, Zeyi Liao, and Xiang Ren. 2022.
\newblock \href {https://doi.org/10.18653/v1/2022.emnlp-main.653} {{R}obust{LR}: A diagnostic benchmark for evaluating logical robustness of deductive reasoners}.
\newblock In \emph{Proceedings of the 2022 Conference on Empirical Methods in Natural Language Processing}, Abu Dhabi, United Arab Emirates. Association for Computational Linguistics.

\bibitem[{Saparov and He(2023)}]{saparov2023language}
Abulhair Saparov and He~He. 2023.
\newblock \href {https://openreview.net/forum?id=qFVVBzXxR2V} {Language models are greedy reasoners: A systematic formal analysis of chain-of-thought}.
\newblock In \emph{The Eleventh International Conference on Learning Representations}.

\bibitem[{Saparov et~al.(2023)Saparov, Pang, Padmakumar, Joshi, Kazemi, Kim, and He}]{saparov2023testing}
Abulhair Saparov, Richard~Yuanzhe Pang, Vishakh Padmakumar, Nitish Joshi, Mehran Kazemi, Najoung Kim, and He~He. 2023.
\newblock \href {https://openreview.net/forum?id=MCVfX7HgPO} {Testing the general deductive reasoning capacity of large language models using {OOD} examples}.
\newblock In \emph{Thirty-seventh Conference on Neural Information Processing Systems}.

\bibitem[{Shao et~al.(2024)Shao, Wang, Zhu, Xu, Song, Bi, Zhang, Zhang, Li, Wu et~al.}]{shao2024deepseekmath}
Zhihong Shao, Peiyi Wang, Qihao Zhu, Runxin Xu, Junxiao Song, Xiao Bi, Haowei Zhang, Mingchuan Zhang, YK~Li, Y~Wu, and 1 others. 2024.
\newblock Deepseekmath: Pushing the limits of mathematical reasoning in open language models.
\newblock \emph{arXiv preprint arXiv:2402.03300}.

\bibitem[{Snell et~al.(2024)Snell, Lee, Xu, and Kumar}]{snell2024scaling}
Charlie Snell, Jaehoon Lee, Kelvin Xu, and Aviral Kumar. 2024.
\newblock Scaling llm test-time compute optimally can be more effective than scaling model parameters.
\newblock \emph{arXiv preprint arXiv:2408.03314}.

\bibitem[{Sun et~al.(2024)Sun, Xu, Liu, Luan, Wang, Shang, Wen, and Yan}]{sun-etal-2024-determlr}
Hongda Sun, Weikai Xu, Wei Liu, Jian Luan, Bin Wang, Shuo Shang, Ji-Rong Wen, and Rui Yan. 2024.
\newblock \href {https://doi.org/10.18653/v1/2024.acl-long.531} {{D}eterm{LR}: Augmenting {LLM}-based logical reasoning from indeterminacy to determinacy}.
\newblock In \emph{Proceedings of the 62nd Annual Meeting of the Association for Computational Linguistics (Volume 1: Long Papers)}, Bangkok, Thailand. Association for Computational Linguistics.

\bibitem[{Tafjord et~al.(2020)Tafjord, Mishra, and Clark}]{tafjord2020proofwriter}
Oyvind Tafjord, Bhavana~Dalvi Mishra, and Peter Clark. 2020.
\newblock Proofwriter: Generating implications, proofs, and abductive statements over natural language.
\newblock \emph{arXiv preprint arXiv:2012.13048}.

\bibitem[{Tian et~al.(2021)Tian, Li, Chen, Xiao, He, and Jin}]{tian-etal-2021-diagnosing}
Jidong Tian, Yitian Li, Wenqing Chen, Liqiang Xiao, Hao He, and Yaohui Jin. 2021.
\newblock \href {https://doi.org/10.18653/v1/2021.emnlp-main.303} {Diagnosing the first-order logical reasoning ability through {L}ogic{NLI}}.
\newblock In \emph{Proceedings of the 2021 Conference on Empirical Methods in Natural Language Processing}, Online and Punta Cana, Dominican Republic. Association for Computational Linguistics.

\bibitem[{Touvron et~al.(2023)Touvron, Lavril, Izacard, Martinet, Lachaux, Lacroix, Rozi{\`e}re, Goyal, Hambro, Azhar et~al.}]{touvron2023llama}
Hugo Touvron, Thibaut Lavril, Gautier Izacard, Xavier Martinet, Marie-Anne Lachaux, Timoth{\'e}e Lacroix, Baptiste Rozi{\`e}re, Naman Goyal, Eric Hambro, Faisal Azhar, and 1 others. 2023.
\newblock Llama: Open and efficient foundation language models.
\newblock \emph{arXiv preprint arXiv:2302.13971}.

\bibitem[{Wang et~al.(2023{\natexlab{a}})Wang, Xu, Lan, Hu, Lan, Lee, and Lim}]{wang2023plan}
Lei Wang, Wanyu Xu, Yihuai Lan, Zhiqiang Hu, Yunshi Lan, Roy Ka-Wei Lee, and Ee-Peng Lim. 2023{\natexlab{a}}.
\newblock Plan-and-solve prompting: Improving zero-shot chain-of-thought reasoning by large language models.
\newblock \emph{arXiv preprint arXiv:2305.04091}.

\bibitem[{Wang et~al.(2025)Wang, Zhao, Wei, and Ren}]{wang2025stepwise}
Siyuan Wang, Enda Zhao, Zhongyu Wei, and Xiang Ren. 2025.
\newblock Stepwise informativeness search for improving llm reasoning.
\newblock \emph{arXiv preprint arXiv:2502.15335}.

\bibitem[{Wang et~al.(2023{\natexlab{b}})Wang, Wei, Schuurmans, Le, Chi, Narang, Chowdhery, and Zhou}]{wang2023selfconsistency}
Xuezhi Wang, Jason Wei, Dale Schuurmans, Quoc~V Le, Ed~H. Chi, Sharan Narang, Aakanksha Chowdhery, and Denny Zhou. 2023{\natexlab{b}}.
\newblock \href {https://openreview.net/forum?id=1PL1NIMMrw} {Self-consistency improves chain of thought reasoning in language models}.
\newblock In \emph{The Eleventh International Conference on Learning Representations}.

\bibitem[{Wei et~al.(2022)Wei, Wang, Schuurmans, Bosma, Xia, Chi, Le, Zhou et~al.}]{wei2022chain}
Jason Wei, Xuezhi Wang, Dale Schuurmans, Maarten Bosma, Fei Xia, Ed~Chi, Quoc~V Le, Denny Zhou, and 1 others. 2022.
\newblock Chain-of-thought prompting elicits reasoning in large language models.
\newblock \emph{Advances in neural information processing systems}, 35:24824--24837.

\bibitem[{xAI(2025)}]{grok3}
xAI. 2025.
\newblock \href {https://x.ai/blog/grok-3/} {Grok 3 beta — the age of reasoning agents}.

\bibitem[{Xie et~al.(2023)Xie, Kawaguchi, Zhao, Zhao, Kan, He, and Xie}]{xie2023self}
Yuxi Xie, Kenji Kawaguchi, Yiran Zhao, James~Xu Zhao, Min-Yen Kan, Junxian He, and Michael Xie. 2023.
\newblock Self-evaluation guided beam search for reasoning.
\newblock \emph{Advances in Neural Information Processing Systems}, 36:41618--41650.

\bibitem[{Xu et~al.(2025)Xu, Lin, Han, Zhao, Liu, and Cambria}]{xu2025large}
Fangzhi Xu, Qika Lin, Jiawei Han, Tianzhe Zhao, Jun Liu, and Erik Cambria. 2025.
\newblock Are large language models really good logical reasoners? a comprehensive evaluation and beyond.
\newblock \emph{IEEE Transactions on Knowledge and Data Engineering}.

\bibitem[{Xu et~al.(2024{\natexlab{a}})Xu, Fei, Luo, Liu, Pan, Wang, Nakov, Lee, and Hsu}]{xu2024aristotle}
Jundong Xu, Hao Fei, Meng Luo, Qian Liu, Liangming Pan, William~Yang Wang, Preslav Nakov, Mong-Li Lee, and Wynne Hsu. 2024{\natexlab{a}}.
\newblock Aristotle: Mastering logical reasoning with a logic-complete decompose-search-resolve framework.
\newblock \emph{arXiv preprint arXiv:2412.16953}.

\bibitem[{Xu et~al.(2024{\natexlab{b}})Xu, Fei, Pan, Liu, Lee, and Hsu}]{xu2024faithful}
Jundong Xu, Hao Fei, Liangming Pan, Qian Liu, Mong-Li Lee, and Wynne Hsu. 2024{\natexlab{b}}.
\newblock Faithful logical reasoning via symbolic chain-of-thought.
\newblock \emph{arXiv preprint arXiv:2405.18357}.

\bibitem[{Yang et~al.(2024)Yang, Yang, Zhang, Hui, Zheng, Yu, Li, Liu, Huang, Wei et~al.}]{yang2024qwen2}
An~Yang, Baosong Yang, Beichen Zhang, Binyuan Hui, Bo~Zheng, Bowen Yu, Chengyuan Li, Dayiheng Liu, Fei Huang, Haoran Wei, and 1 others. 2024.
\newblock Qwen2. 5 technical report.
\newblock \emph{arXiv preprint arXiv:2412.15115}.

\bibitem[{Yao et~al.(2023)Yao, Yu, Zhao, Shafran, Griffiths, Cao, and Narasimhan}]{yao2023tree}
Shunyu Yao, Dian Yu, Jeffrey Zhao, Izhak Shafran, Tom Griffiths, Yuan Cao, and Karthik Narasimhan. 2023.
\newblock Tree of thoughts: Deliberate problem solving with large language models.
\newblock \emph{Advances in neural information processing systems}, 36:11809--11822.

\bibitem[{Yeo et~al.(2025)Yeo, Tong, Niu, Neubig, and Yue}]{yeo2025demystifying}
Edward Yeo, Yuxuan Tong, Morry Niu, Graham Neubig, and Xiang Yue. 2025.
\newblock Demystifying long chain-of-thought reasoning in llms.
\newblock \emph{arXiv preprint arXiv:2502.03373}.

\bibitem[{Zhang et~al.(2025)Zhang, Yang, Yuan, and Yao}]{zhang2025cumulative}
Yifan Zhang, Jingqin Yang, Yang Yuan, and Andrew~C Yao. 2025.
\newblock \href {https://openreview.net/forum?id=grW15p4eq2} {Cumulative reasoning with large language models}.
\newblock \emph{Transactions on Machine Learning Research}.

\bibitem[{Zhang et~al.(2023)Zhang, Zhang, Li, and Smola}]{zhang2023automatic}
Zhuosheng Zhang, Aston Zhang, Mu~Li, and Alex Smola. 2023.
\newblock \href {https://openreview.net/forum?id=5NTt8GFjUHkr} {Automatic chain of thought prompting in large language models}.
\newblock In \emph{The Eleventh International Conference on Learning Representations}.

\bibitem[{Zhou et~al.(2023)Zhou, Sch{\"a}rli, Hou, Wei, Scales, Wang, Schuurmans, Cui, Bousquet, Le, and Chi}]{zhou2023leasttomost}
Denny Zhou, Nathanael Sch{\"a}rli, Le~Hou, Jason Wei, Nathan Scales, Xuezhi Wang, Dale Schuurmans, Claire Cui, Olivier Bousquet, Quoc~V Le, and Ed~H. Chi. 2023.
\newblock \href {https://openreview.net/forum?id=WZH7099tgfM} {Least-to-most prompting enables complex reasoning in large language models}.
\newblock In \emph{The Eleventh International Conference on Learning Representations}.

\end{thebibliography}

\clearpage

\appendix

\section{Computation for Semantic Overlap and Cumulative Connectivity}
\label{compute}

For each fact \( f \), rule \( r \), as well as the hypothesis \( \mathcal{H} \), we extract a structured triple:

\vspace{3pt}
\begin{center}
\begin{tabular}{ll}
    \( (\mathcal{S}_{\text{Subj}}^{f}, \mathcal{S}_{\text{Pred}}^{f}, \mathcal{S}_{\text{SP}}^{f}) \) & \text{from a fact,} \\[1.5ex]
    \( (\mathcal{S}_{\text{Subj}}^{r}, \mathcal{S}_{\text{Pred}}^{r}, \mathcal{S}_{\text{SP}}^{r}) \) & \text{from a rule,} \\[1.5ex]
    \( (\mathcal{S}_{\text{Subj}}^{\mathcal{H}}, \mathcal{S}_{\text{Pred}}^{\mathcal{H}}, \mathcal{S}_{\text{SP}}^{\mathcal{H}}) \) & \text{from the hypothesis.}
\end{tabular}
\end{center}
\vspace{3pt}

\noindent In each triple:
\begin{itemize}[noitemsep,nolistsep]
    \item \( \mathcal{S}_{\text{Subj}} \) denotes the \textit{Set of Subjects},
    \item \( \mathcal{S}_{\text{Pred}} \) denotes the \textit{Set of Predicates},
    \item \( \mathcal{S}_{\text{SP}} \) denotes the \textit{Set of Subject-Predicate pairs} identified via parent-child relations in \texttt{spaCy}.
\end{itemize}
\vspace{2pt}
\noindent These sets are extracted using \texttt{spaCy}'s dependency parser. We represent them as sets to account for the possibility of multiple subjects and predicates within a single fact, rule, or hypothesis.

The \textbf{semantic overlap} \( Sem(f, r) \) between a fact \( f \) and a rule \( r \) is defined as:
\begin{align}
Sem(f, r) =\ & 0.25 \cdot \mathbb{I}(\mathcal{S}_{\text{Subj}}^{f} \cap \mathcal{S}_{\text{Subj}}^{r} \neq \emptyset) \nonumber \\
& +\ 0.25 \cdot \mathbb{I}(\mathcal{S}_{\text{Pred}}^{f} \cap \mathcal{S}_{\text{Pred}}^{r} \neq \emptyset) \nonumber \\
& +\ 0.5 \cdot \mathbb{I}(\mathcal{S}_{\text{SP}}^{f} \cap \mathcal{S}_{\text{SP}}^{r}) \neq \emptyset),
\end{align}

\noindent and similarly, the \textbf{semantic overlap} \( Sem(r, \mathcal{H}) \) between a rule \( r \) and hypothesis \( \mathcal{H} \) is defined as:
\begin{align}
Sem(r, \mathcal{H}) =\ & 0.25 \cdot \mathbb{I}(\mathcal{S}_{\text{Subj}}^{r} \cap \mathcal{S}_{\text{Subj}}^{\mathcal{H}} \neq \emptyset) \nonumber\\
                   & +\ 0.25 \cdot \mathbb{I}(\mathcal{S}_{\text{Pred}}^{r} \cap \mathcal{S}_{\text{Pred}}^{\mathcal{H}} \neq \emptyset) \nonumber\\
                   & +\ 0.5 \cdot \mathbb{I}(\mathcal{S}_{\text{SP}}^{r} \cap \mathcal{S}_{\text{SP}}^{\mathcal{H}} \neq \emptyset),
\end{align}

\noindent where \( \mathbb{I}(\cdot) \) is an indicator function:
\[
\mathbb{I}(\text{condition}) =
\begin{cases} 
1, & \text{if condition is true} \\ 
0, & \text{otherwise}
\end{cases}
\]

\noindent We set the coefficients as $0.25$, $0.25$, and $0.5$ respectively, such that the semantic overlap is upper-bounded by $1$, achieved when all three conditions are satisfied. Partial (i.e, subject or predicate) overlaps are assigned with non-zero coefficient ($0.25$) because they may still indicate logical relevance. For example, in the case where the fact is \textit{“Dave is hungry.”} and the rule is \textit{“If someone is hungry, they are uncomfortable.”}, only the predicate overlaps, but the fact is logically connected to the rule.

The \textbf{cumulative connectivity} \( \mathcal{C}(f, \mathcal{R}) \) between a fact \( f \) and the entire rule set $\mathcal{R}$ is defined as the sum of its semantic overlap with each rule in $\mathcal{R}$, i.e.,

\begin{align}
\mathcal{C}(f, \mathcal{R}) = \sum_{r \in \mathcal{R}} Sem(f, r).
\end{align}

A higher cumulative connectivity value indicates that the fact \( f \) is likely to initiate more reasoning branches through its relevant rules.

\section{Linear Premise Search in LogicTree}
\label{linear}
In our framework, premise search is simplified by decomposing it into forward (rule) selection and backward (fact) selection (\S~\ref{3.2}), resulting in a linear rather than combinatorial search process.

During forward selection, the framework takes a fact as an anchor and identifies all relevant rules. Although multiple rules may be retrieved in a single LLM query, LLM can perform a process analogous to a linear iteration over the rule set, evaluating each rule independently for relevance without requiring joint combinations.

Similarly, in backward selection, we consider a general conjunctive reasoning case (e.g., $f_1 \land f_2 \land ... \land f_n\land r_1 \Rightarrow\ d_1$), where an anchor fact $f_1$ partially satisfies a rule $r_1$. Once this rule $r_1$ is identified, the remaining required facts ($f_2$ through $f_n$) are identified from the rule’s conditions and subsequently checked for existence in Fact Repository by Backward Selection Module. LLM can implement this step in a way that resembles a linear scan by verifying the existence of each required fact individually.

In contrast, CoT and ToT require combinatorial search for fact–rule combinations, where the facts and rules must be jointly selected and logically relevant to each other as shown in Table~\ref{table_complexity}, thereby increasing the complexity of LLM’s search process.

\section{Experimental Details}
\subsection{Dataset}
\label{dataset}
We evaluate on five English-language logical reasoning datasets, as detailed below:

\noindent\textbf{RobustLR} ~\citep{sanyal-etal-2022-robustlr} includes Logical Contrast and Logical Equivalence sets for testing the logical robustness on conjunctive, disjunctive, and contrapositive reasoning. We randomly sample 240 examples from the
test set.

\noindent\textbf{PrOntoQA-OOD} ~\citep{saparov2023testing} is a synthetic question-answering dataset using fictional names. For evaluation, we use the most challenging 4-hop subset. We randomly sample 200 examples from the test set.

\noindent\textbf{ProofWriter} ~\citep{tafjord2020proofwriter} is a commonly used benchmark for deductive logical reasoning. We evaluate the open-world assumption (OWA) subset, focusing on the hardest depth-5 subset. We randomly sample 600 examples from the test set.

\noindent\textbf{ParaRules} ~\citep{ijcai2020p0537} paraphrases data from ProofWriter into more natural language using crowdsourcing, enhancing text diversity and naturalness. We randomly sample 600 examples from the test set.

\noindent\textbf{LogicNLI} ~\citep{tian-etal-2021-diagnosing} is the most challenging dataset, featuring a large premise space and numerous reasoning paths, only one of which leads to the proof. We randomly sample 150 examples from the
test set.

Few-shot demonstrations for each LLM module are sampled from the training set of each dataset. An example of each dataset is shown in Appendix~\ref{example_and_prompt}.

\subsection{Models}
\label{model}

Here are the versions of OpenAI's models:

\noindent GPT-4o-mini: \texttt{gpt-4o-mini-2024-07-18}

\noindent GPT-4o: \texttt{gpt-4o-2024-08-06}

\noindent o1-mini: \texttt{o1-mini-2024-09-12}

\noindent o3-mini (medium): \texttt{o3-mini-2025-01-31}

All OpenAI models are accessed through OpenAI API\footnote{\url{https://platform.openai.com/docs/overview}}. 
Llama-3.3 70B is accessed through Together AI API\footnote{\url{https://www.together.ai/models/llama-3-3-70b}}.
We set the temperature to 0.1 for all experiments to encourage more deterministic generation. All results are obtained from a single run. We utilize the Microsoft Guidance library in our implementation\footnote{\url{https://github.com/guidance-ai/guidance}}.

The version of \texttt{spaCy} model used in our framework is \texttt{en\_core\_web\_lg-3.8.0} (382 MB).

\section{Error Analysis on CoT, o1-mini, o3-mini, LogicTree}
\label{error}
We manually conduct error analysis on CoT, o1-mini, o3-mini, and our framework using ProofWriter dataset. For CoT, we randomly sample 100 failed proofs, while for o1-mini, o3-mini, and our framework, we analyze all failed cases. We categorize the errors into three types: (1) insufficient exploration, (2) wrong derivation, and (3) hallucinated premise. The proportion of these error types for each method, along with illustrative examples, is shown in Figure~\ref{error_anals}. Our framework exhibits significantly fewer errors caused by insufficient exploration. In addition, our framework does not suffer from hallucinated premises, as access to the hypothesis is restricted to Verification Module only. This prevents the generation of unsupported premises that favor verifying the hypothesis during premise selection and inference.

Our findings on why CoT struggles with complex logical reasoning align with prior research: (1) it faces difficulty when premises are unordered and contain distractions ~\citep{pmlr-v235-chen24i}, and (2) it lacks systematic exploration when reasoning requires navigating extensive branching ~\citep{saparov2023language}. The high branching factor that complicates exploration, along with sensitivity to distractions, also limits the performance of o1-mini and o3-mini in complex logical reasoning compared to their effectiveness in coding and math. To address this, a modular method for precise premise selection which strengthens robustness to distractions, combined with an algorithm-guided approach for systematic proof searching, provides a promising foundation. Our framework builds upon and extends these components, addressing their limitations to develop a logically complete algorithm that enables rigorous and coherent reasoning, ultimately achieving superior proof accuracy.

\section{Number of Reasoning Steps, Generated Tokens, and Inference Time}
\label{steps}
The following elaborates on how we measure the number of reasoning steps for each approach.

\noindent (1) For CoT, we define one reasoning step as a combination of premise selection and an inference based on the selected premises. To make step counting explicit in LLM's output, we number each reasoning step in few-shot demonstrations (Figure~\ref{cot_prompt}). 

\noindent (2) For SI, we set the maximum number of iterations to 10, as we find the framework typically fails to generate new derivations beyond this point. Each iteration consists of one query to LLM selection module and one query to LLM inference module. The process terminates early if the hypothesis is successfully verified. We define the total number of reasoning steps in SI as the total number of LLM module queries made across the iterations.

\noindent (3) For CR, we use the framework’s default hyper-parameters for reasoning. The total number of reasoning steps in CR is calculated as the total number of LLM module queries made during the iterations, plus the number of steps in the final CoT reasoning process.

\noindent (4) For ToT, LAMBADA, and LogicTree, each query to an LLM module is counted as one reasoning step. We set the step limit $L$ (in Algorithm~\ref{algo1}) to 80 for all methods. Tree search terminates early if the hypothesis is successfully verified.

\noindent (5) Since reasoning with o1-mini and o3-mini includes unobservable intermediate steps, we exclude them from the analysis of reasoning steps.

Table~\ref{table_reason_step} shows the average number of reasoning steps across five logical reasoning datasets for different methods. Table~\ref{output_token_time} presents the average number of generated tokens and inference time for those different methods. The number of generated tokens is obtained using \texttt{completion\_tokens} from the completion response.

\section{Performance Analysis: Forward vs. Backward Reasoning}
\label{for_vs_back}
In Table~\ref{table_forward_backward}, we present in-depth analysis to explain why backward reasoning requires more reasoning steps than forward reasoning in iterative tree search, based on three key factors: (1) utilization of historical knowledge; (2) evaluation of hypothesis and its negation; and (3) branching complexity in reasoning paths. 

The lower proof accuracy of LAMBADA (backward reasoning) compared to forward reasoning methods (ToT and LogicTree) can be explained by two main reasons: (1) LLMs demonstrate higher stepwise inference accuracy in forward reasoning as shown in Figure~\ref{fig3}b (\S~\ref{5.1}); (2) the larger number of reasoning steps in backward reasoning increases the likelihood of making errors.

\section{Using LLM Modules for Premise Prioritization}
\label{llm_for_prior}
The prompts used for the LLM modules in fact and rule ranking are provided in Appendix~\ref{example_and_prompt}. We do not apply predefined criteria in these prompts, allowing us to assess LLM's inherent ability on premise prioritization. For the average reasoning steps (\textit{LLM-based prioritize}) reported in Table~\ref{table2}, we do not include LLM queries related to fact and rule ranking. Even without this overhead, LLM-based premise prioritization results in more reasoning steps than our LLM-free heuristics. This reflects the limitation in LLM-based proof planning. Successfully guiding the reasoning process requires the model to already have an accurate understanding of how to reach the proof in advance, yet this ability is not evidenced by the limited performance of CoT (Table~\ref{table1}). Based on the results in Table~\ref{table2}, our simple LLM-free heuristics prove effective for strategic proof exploration.

\section{Extension to Mathematical Reasoning}
\label{math_extension}
LogicTree is primarily designed to strengthen the logical reasoning capabilities of LLMs. Beyond its original focus, it can be readily adapted to other types of reasoning tasks that start from a given set of information. Our framework systematically combines relevant information for derivation, leverages derived information, and facilitates structured problem solving. In Figure~\ref{math_demo}, we illustrate how LogicTree performs mathematical reasoning on an example from GSM8K~\citep{cobbe2021training}. We further evaluate and compare the accuracy of CoT, ToT, and LogicTree on 300 randomly sampled examples from  GSM8K test set and 300 randomly sampled examples from MathQA~\citep{amini-etal-2019-mathqa} test set. All methods are evaluated on Qwen2.5-7B~\citep{yang2024qwen2}, with LogicTree demonstrating superior performance.

\section{Dataset Example and Prompt for LLM Modules}
\label{example_and_prompt}
Figure~\ref{logicnli}, Figure~\ref{prontoqa}, Figure~\ref{proofwriter}, Figure~\ref{RobustLR}, Figure~\ref{ParaRules} show an example of LogicNLI, PrOntoQA-OOD, ProofWriter, RobustLR, ParaRules, respectively.

Figure~\ref{o1_o3_prompt} shows the prompts for reasoning with OpenAI's o1-mini and o3-mini model. We use the same prompts for all the five datasets.

Figure~\ref{cot_prompt} shows the prompts for chain-of-thought. The instructions in the prompt are identical across all five datasets, while the demonstrations are sampled from the training set of each respective  dataset. We use examples from ProofWriter as illustrations.

Figure~\ref{forward_prompt}, Figure~\ref{backward_prompt}, Figure~\ref{derive_prompt}, Figure~\ref{verify_prompt} separately show the prompts for Forward Selection Module, Backward Selection Module, Derivation Module, Verification Module in our framework. The instructions in the prompt are consistent across all five datasets, while the demonstrations are sampled from the training set of each respective dataset. We use examples from ProofWriter as illustrations.

Figure~\ref{fact_rank_prompt} and Figure~\ref{rule_rank_prompt} respectively show the prompts for
Fact Ranking Module and Rule Ranking Module, which are used in the ablation studies with results presented in Table~\ref{table2}.

%%%%%%%%%%%%%%%%%%%%%%%%%%%%%%%%%%%%%%
\begin{figure*}[t]
\centering
\includegraphics[width=1\textwidth]{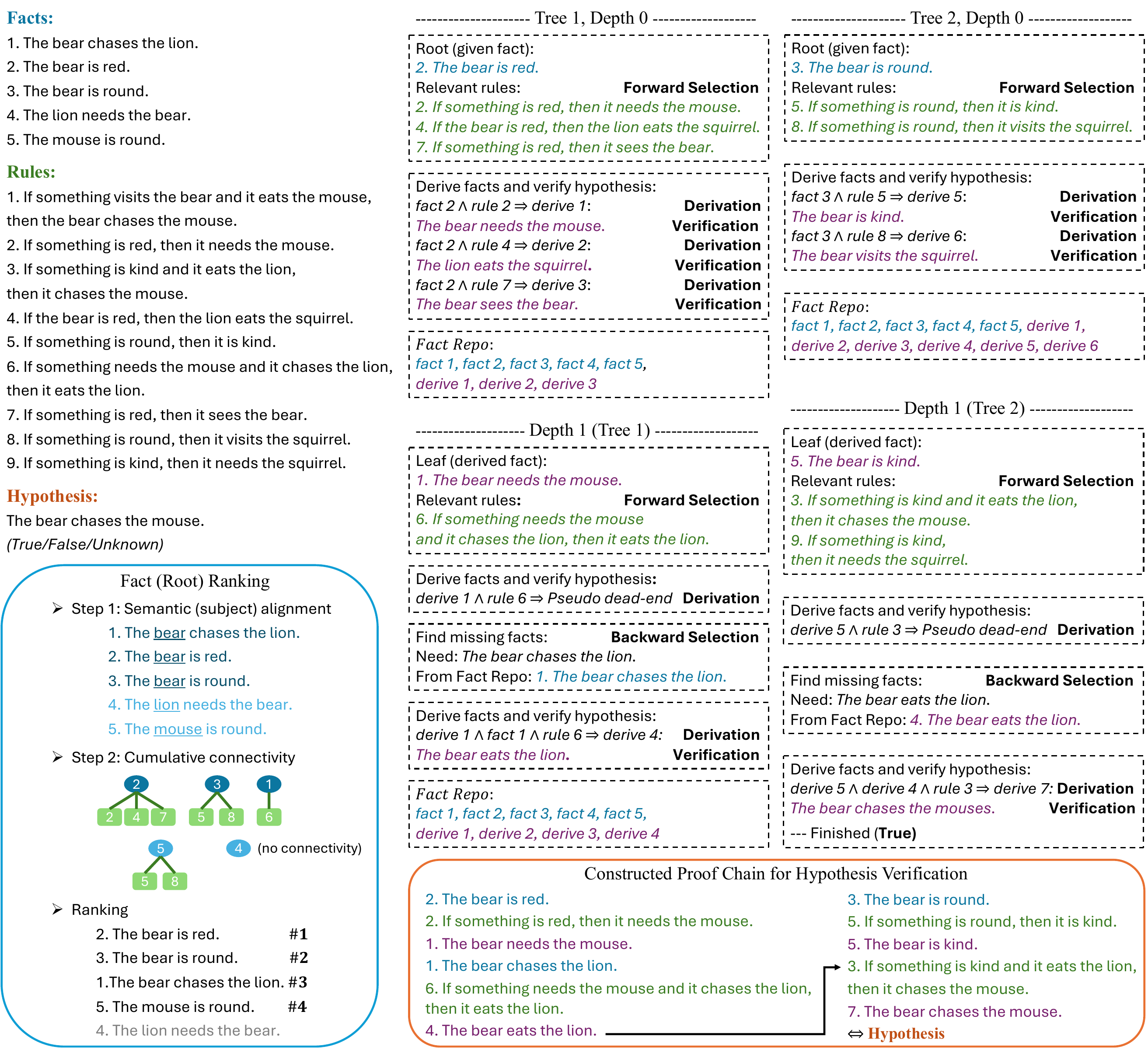}
\caption{The proof exploration trace of LogicTree on an example from ProofWriter ~\citep{tafjord2020proofwriter}. Rule Ranking, Derivation HashMap and some leaves (\textit{derive2, derive3, derive6}) are omitted for brevity.}
\label{fig5}
\end{figure*}
%%%%%%%%%%%%%%%%%%%%%%%%%%%%%%%%%%%%%%

%%%%%%%%%%%%%%%%%%%%%%%%%%%%%%%%%%%%%%
\begin{figure*}[t]
\centering
\includegraphics[width=1\textwidth]{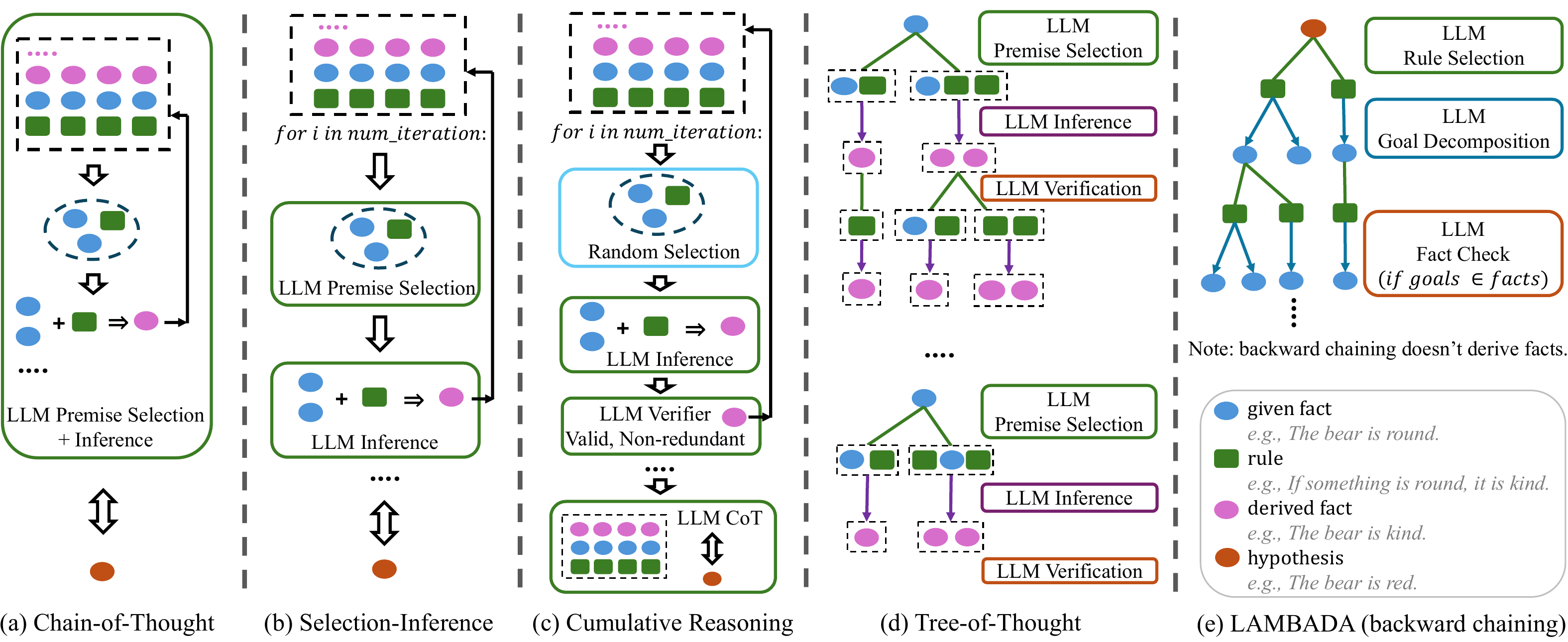}
\caption{Schematic illustrations of baseline approaches for logical reasoning with LLMs: (a) Chain-of-Thought; (b) Selection-Inference; (c) Cumulative Reasoning; (d) Tree-of-Thought; (e) LAMBADA.}
\label{fig6}
\end{figure*}
%%%%%%%%%%%%%%%%%%%%%%%%%%%%%%%%%%%%%%
\clearpage

%%%%%%%%%%%%%%%%%%%%%%%%%%%%%%%%%%%%%%%%%%%%%%%%%
\begin{figure*}[htb]
  \centering
  \small
  \begin{minipage}[t]{.6\linewidth}
\begin{algorithm}[H]
\caption{LogicTree DFS Algorithm}\label{algo1}
\begin{algorithmic}[1]
\Statex \hspace{-1.5em} \textbf{Require:} 
\Statex \hspace{-1.5em} Premises $\mathcal{F} \cup \mathcal{R}$, Hypothesis $\mathcal{H}$
\Statex \hspace{-1.5em} Large language model $LLM$, NLP library \texttt{spaCy}

    \vspace{1.2mm}
    \State $s \gets 0$ \textcolor{gray}{\small $\triangleright$ Number of reasoning steps}
    \State $step\_limit \gets L$
    \For {each $f_i \in \mathcal{F}$}
        \State $result \gets \texttt{Early\_Stop}(LLM, f_i, \mathcal{H}, s)$
        \If{$result \neq$ Unknown}
            \State \Return $result$ \textcolor{gray}{\small $\triangleright$ Verified}
        \EndIf
    \EndFor 
    \vspace{0.8mm}

    \State $Fact\_Repo \gets \mathcal{F}$
    \State $Derivation\_HashMap \gets \{\}$
    \State $\mathcal{F}_{rank} \mathrel{\gets} \textcolor{deepblue}{\texttt{Fact\_Ranking}}(\texttt{spaCy}, \mathcal{F}, \mathcal{R})$
    \For{each $f_i \in \mathcal{F}_{rank}$}
        \If{$s > step\_limit$} 
            \State \textbf{break}
        \EndIf
        \State $Tree \gets [f_i]$ \textcolor{gray}{\small $\triangleright$ Initiate the root of a tree}
        \While{$Tree \neq \emptyset$}
            \State $f_{l} \gets Tree.\texttt{pop()}$ \textcolor{gray}{\small $\triangleright$ LIFO, get the leaf fact}
            \State $\mathcal{R}^{relv} \gets \textcolor{darkorange}{\texttt{Forward\_Selection}}($
            \Statex \hspace{4.8cm} $LLM, f_{l}, \mathcal{R}, s+1)$ 
            \State $\mathcal{R}^{relv}_{rank} \gets \textcolor{deepblue}{\texttt{Rule\_Ranking}}(\texttt{spaCy}, \mathcal{R}^{relv}, \mathcal{H})$ 
            \For{each $r_i \in \mathcal{R}^{relv}_{rank}$} % \textcolor{gray}{\small $\triangleright$ Extend branches if $\mathcal{R}^{relv} \neq \emptyset$}
                \State $d_i \gets \texttt{Inference}(LLM, f_{l}, r_i, Fact\_Repo, s)$
                \State $result \gets \texttt{Early\_Stop}(LLM, d_i, \mathcal{H}, s)$
                \If{$result \neq$ Unknown}
                    \State \Return $result$ \textcolor{gray}{\small $\triangleright$ Verified}
                \EndIf
                \If{$(d_i \neq \text{null}) \land (d_i \notin Fact\_Repo)$}
                    \State $Tree.\texttt{append}(d_i)$ \textcolor{gray}{\small $\triangleright$ Extend branches}
                    \State  $Fact\_Repo\gets Fact\_Repo \cup \{d_i\}$
                    \State $Derivation\_HashMap[d_i] \gets path_{d_{i}}$
                \EndIf

            \EndFor
        \EndWhile
    \EndFor
    \State \Return \text{Unknown}
\end{algorithmic}
  \end{algorithm}
  \end{minipage}
  \hfill
  \begin{minipage}[t]{.38\linewidth}
\begin{algorithm}[H]
\caption{Inference Function}\label{algo2}
\begin{algorithmic}[1]
\Statex \hspace{-1.5em} \textbf{Require:} 
\Statex \hspace{-1.5em} $f_{l}, r_i, Fact\_Repo, s$
\Statex \hspace{-1.5em} Large language model $LLM$

\vspace{1.2mm}
\State $d_i \gets \textcolor{darkorange}{\texttt{Derivation}}($
\Statex \hspace{2.4cm} $LLM, f_l, r_i, s+1)$
\If{$d_i \neq \text{null}$}
    \State \Return $d_i$
\EndIf

\State \textcolor{gray}{\small $\triangleright$ Use textual pattern matching}
\If{$\texttt{PseudoDeadEnd}(d_i)$}
    \State \textcolor{gray}{\small $\triangleright$ Search missing fact}
    \State $f_s \gets \textcolor{darkorange}{\texttt{Backward\_Selection}}($
    \Statex \hspace{0.3cm} $LLM, f_l, r_i, Fact\_Repo, s+1)$
    \If{$f_s \neq \text{null}$}
    \State $d_i^{\prime} \gets \textcolor{darkorange}{\texttt{Derivation}}($
    \Statex \hspace{1.8cm} $LLM, f_l \land f_s, r_i, s+1)$
    \State \Return $d_i^{\prime}$
    \EndIf
\EndIf
\State \Return \text{null}
\end{algorithmic}
\end{algorithm}

\vspace{1.8cm} % Adjust spacing
\begin{algorithm}[H]
\caption{Early\_Stop Function}\label{algo3}
\begin{algorithmic}[1]
\Statex \hspace{-1.5em} \textbf{Require:} 
\Statex \hspace{-1.5em} $f_i$ or $d_i$, $\mathcal{H}$, $s$
\Statex \hspace{-1.5em} Large language model $LLM$

\vspace{1.2mm}
\State $result \gets \textcolor{darkorange}{\texttt{Verification}}($
\Statex \hspace{1.4cm} $LLM, f_i\text{ or }d_i, \mathcal{H}, s+1)$
\If{$result \neq \text{Unknown}$}
    \State \textcolor{gray}{\small $\triangleright$ $\mathcal{H}$ is verified}
    \State \textcolor{gray}{\small $\triangleright$ $result$ is either \textit{True} or \textit{False}}
    \State \Return $result$
\EndIf
\State \textcolor{gray}{\small $\triangleright$ $\mathcal{H}$ is not verified}
\State \Return Unknown

\end{algorithmic}
\end{algorithm}

  \end{minipage}

\end{figure*}
%%%%%%%%%%%%%%%%%%%%%%%%%%%%%%%%%%%%%%%%%%%%%%%%%

%%%%%%%%%%%%%%%%%%%%%%%%%%%%%%%%%%%%%%%%%%%%%%%%%
\begin{table*}[ht]
\centering
\small
\begin{tabular}{
@{}>{\raggedright\arraybackslash}p{.45\textwidth}
@{\hspace{4.5em}}
>{\raggedright\arraybackslash}p{.45\textwidth}@{}
}
\toprule
\multicolumn{2}{@{}l@{}}{
\parbox[t]{\textwidth}{
\textbf{Facts $\mathcal{F}$ + Rules $\mathcal{R}$:} \\[0.4em]
{[...]} \textcolor{deepblue}{The bald eagle likes the dog.} The bald eagle needs the tiger. The bald eagle sees the tiger. \textcolor{deepblue}{The bald eagle needs the dog.} \\ The dog is blue. The dog sees the tiger. The rabbit is green. The tiger needs the bald eagle. {[...]}\\
{[...]} \\ If someone needs the bald eagle and the bald eagle sees the tiger then they are rough. \\
\textcolor{deepgreen}{If someone needs the dog and they like the dog then they like the tiger.} \\
If someone likes the bald eagle then the bald eagle needs the dog. \\
If someone is rough and they like the dog then the dog needs the tiger.\\
{[...]}\\[0.6em]
\textbf{Hypothesis:} \\[0.4em]
\textcolor{darkorange}{The bald eagle likes the tiger.}\\[-0.6em]
}
} \\ 
\midrule 
% ---------- Row 1 ----------
% ---------- CoT ----------
\begin{minipage}[t][4.5cm][b]{.45\textwidth}
\textbf{$\blacktriangleright$ CoT} \\ [0.2em]
\# Premise Search \\ [0.2em]
\textcolor{deepblue}{The bald eagle likes the dog.} \\ [0.2em]
\textcolor{deepblue}{The bald eagle needs the dog.} \\ [0.2em]
\textcolor{deepgreen}{If someone needs the dog and they like the dog then they like the tiger.} \\
\vfill
One-step search \\ [0.2em]
\textbf{\textit{Combinatorial complexity}} \\
\end{minipage}
&
% ---------- ToT ----------
\begin{minipage}[t][4.5cm][b]{.45\textwidth}
\textbf{$\blacktriangleright$ ToT} \\ [0.2em]
\# Parent node (anchor) \\ [0.2em]
\faMapMarker \hspace{0.15em} \textcolor{deepblue}{The bald eagle likes the dog.} \\ [0.2em]
\# Premise Search \\ [0.2em]
\textcolor{deepblue}{The bald eagle needs the dog.} \\ [0.2em]
\textcolor{deepgreen}{If someone needs the dog and they like the dog then they like the tiger.} \\
\vfill
One-step search \\ [0.2em]
\textbf{\textit{Combinatorial complexity}} \\
\end{minipage}
\\
\hdashline
% ---------- Row 2 ----------
% ---------- lambada ----------
\begin{minipage}[t][7cm][b]{.45\textwidth}
\textbf{$\blacktriangleright$ LAMBADA} \\ [0.2em]
\# Parent node (anchor) \\ [0.2em]
\faMapMarker \hspace{0.2em}\textcolor{darkorange}{The bald eagle likes the tiger.} \\ [0.2em]
\# Premise (rule) Search \\ [0.2em]
\textcolor{deepgreen}{If someone needs the dog and they like the dog then they like the tiger.} \\ [0.2em]
\# Fact Check (if $f \in \mathcal{F}$)\\ [0.2em]
\textcolor{deepblue}{The bald eagle needs the dog.} \\ [0.2em]
\textcolor{deepblue}{The bald eagle likes the dog.} \\
\vfill
One-step search + one-step check \\ [0.2em]
\textbf{\textit{Linear complexity}} \\ [-0.6em]
\end{minipage}
&
% ---------- LogicTree ----------
\begin{minipage}[t][7cm][b]{.45\textwidth}
\textbf{$\blacktriangleright$ LogicTree} \\ [0.2em]
\# Parent node (anchor) \\ [0.2em]
\faMapMarker \hspace{0.15em} \textcolor{deepblue}{The bald eagle likes the dog.} \\ [0.2em]
\# Forward (rule) Selection \\ [0.2em]
\textcolor{deepgreen}{If someone needs the dog and they like the dog then they like the tiger.} \\ [0.2em]
% \# Pseudo dead-end \\ [0.2em]
\# fact-rule pair (anchor) \\ [0.2em]
\faMapMarker \hspace{0.15em} \textcolor{lightblue}{The bald eagle likes the dog.} + \textcolor{lightgreen}{If someone needs the dog and they like the dog then they like the tiger.}\\ [0.2em]
\# Backward (fact) Selection \\ [0.2em]
\textcolor{deepblue}{The bald eagle needs the dog.} \\
\vfill
Two-step search \\ [0.2em]
\textbf{\textit{Linear complexity}} \\ [-0.6em]
\end{minipage} \\
\bottomrule
\end{tabular}
\caption{Complexity of premise search in CoT, ToT, LAMBADA, and LogicTree. An example involving \textit{conjunctive reasoning} from ProofWriter is used for analysis. Note that although ToT's complexity becomes linear in non-conjunctive cases, the reasoning type is not known beforehand. Therefore, ToT must retain its higher search complexity in the general case to accommodate complex scenarios.}
\label{table_complexity}
\end{table*}
%%%%%%%%%%%%%%%%%%%%%%%%%%%%%%%%%%%%%%%%%%%%%%%%%

%%%%%%%%%%%%%%%%%%%%%%%%%%%%%%%%%%%%%%%%%%%%%%%%%
\begin{table*}[t]
\centering
\small 
\renewcommand{\arraystretch}{1.2}
\begin{tabular}{@{}p{4.3cm} p{5.4cm} p{5.4cm}@{}}
\toprule
\textbf{Factor} & \textbf{Forward Reasoning} & \textbf{Backward Reasoning} \\
\midrule

\textbf{Utilization of historical knowledge (intermediate derivations)} & 
(i) \textit{Immediate utilization:} Derived facts are immediately valid and can be utilized across reasoning chains through caching, enabling efficient information sharing and minimizing redundant computation. \newline

(ii) \textit{Dead ends still contribute:} Even if a reasoning chain fails to reach the final hypothesis, its intermediate derived facts are valid and can be utilized to support other inference paths. & 

(i) \textit{Delayed utilization:}  Intermediate facts are only validated upon completing a successful reasoning chain and cannot be utilized across other chains while the current chain is still in progress. \newline

(ii) \textit{Dead ends yield nothing:} If a reasoning chain fails, none of its intermediate facts are proven valid and therefore cannot be utilized in other inference paths.\\ \\

\textbf{Evaluation of hypothesis and its negation} & 
Forward reasoning starts from the given premises and derives logically valid conclusions, using those results to determine the truth value of the hypothesis. The derivation process does not explicitly attempt to prove or disprove the hypothesis.
& 
Backward reasoning must start from and evaluate both the hypothesis and its negation to determine the truth value, as failure to prove one does not imply the truth of the other—due to the possibility of an \textit{unknown} outcome. This requirement increases the overall reasoning steps. \newline 
E.g., to evaluate the hypothesis \textcolor{darkorange}{Dave is blue}, backward reasoning explores both the supporting path starting with \newline \textcolor{deepgreen}{If ..., then Dave is \underline{blue}},\newline
and the opposing path starting with \newline
\textcolor{deepgreen}{If ..., then Dave is \underline{not blue}}. \\ \\

\textbf{Branching complexity in reasoning paths} & 
Forward reasoning applies rules to known facts independently. \newline \newline
Path count grows \textbf{additively}. \newline 
E.g., given the facts:\newline
\textcolor{deepblue}{Dave is blue} and \textcolor{deepblue}{Dave is cold}, \newline
suppose there are: \newline
$m$ rules of the form: \textcolor{deepgreen}{If Dave is blue, then ...} \newline
$n$ rules of the form: \textcolor{deepgreen}{If Dave is cold, then ...} The total number of paths to be explored is $m + n$.& 
Backward reasoning explores all combinations of rules whose conclusions match intermediate goals. \newline Path count grows \textbf{multiplicatively}. \newline
E.g., to satisfy the conjunction of goals: \newline
\textcolor{deepblue}{Dave is blue} and \textcolor{deepblue}{Dave is cold}, \newline 
suppose there are: \newline
$m$ rules of the form: \textcolor{deepgreen}{If ..., then Dave is blue.} \newline
$n$ rules of the form: \textcolor{deepgreen}{If ..., then Dave is cold.} The total number of paths to be explored is $m * n$. \\

\bottomrule
\end{tabular}
\caption{Analysis of three key factors that lead to more \textbf{reasoning steps} in backward reasoning compared to forward reasoning in \textit{iterative tree search} for logical reasoning tasks: (1) Utilization of historical knowledge, (2) Evaluation of hypothesis and its negation, and (3) Branching complexity in reasoning paths. Together, these factors result in a higher average number of reasoning steps in backward reasoning, as illustrated in Figure~\ref{fig4} and Figure~\ref{fig8}.}
\label{table_forward_backward}
\end{table*}
%%%%%%%%%%%%%%%%%%%%%%%%%%%%%%%%%%%%%%%%%%%%%%%%%

%%%%%%%%%%%%%%%%%%%%%%%%%%%%%%%%%%%%%%%%%%%%%%%%%
\begin{table*}
\small
\begin{center}
\setlength{\tabcolsep}{12.5pt}
\renewcommand{\arraystretch}{1.2}
\begin{tabular}{lccccc}
\toprule
  & LogicNLI & ParaRules & PrOntoQA-OOD & ProofWriter & RobustLR \\
\midrule
\multicolumn{5}{l}{$\bullet$ \textbf{\emph{Llama-3.3 70B}}} \\
% \addlinespace[1pt]

LogicTree w/o fact repo & 68.7 & 81.7 & 90.0 & 88.7 & 92.1 \\
\rowcolor{gray!20}  \bf LogicTree & \textbf{74.7} & \textbf{92.3} & \textbf{97.0} & \textbf{95.8} & \textbf{97.5}\\

\midrule

\multicolumn{5}{l}{$\bullet$ \textbf{\emph{GPT-4o}}} \\
% \addlinespace[1pt]

LogicTree w/o fact repo & 72.0 & 87.5 & 88.0 & 90.0 & 93.8 \\
\rowcolor{gray!20}  \bf LogicTree & \textbf{78.7} & \textbf{96.3} & \textbf{99.0} & \textbf{97.0} & \textbf{97.9}\\

\bottomrule
\end{tabular}

\end{center}
\caption{
Impact of the fact repository (fact repo) on proof accuracy across five datasets within our framework.
}
\label{table_fact_repo}
\end{table*}
%%%%%%%%%%%%%%%%%%%%%%%%%%%%%%%%%%%%%%%%%%%%%%%%%

%%%%%%%%%%%%%%%%%%%%%%%%%%%%%%%%%%%%%%
\begin{figure*}[t]
\centering
\includegraphics[width=1\textwidth]{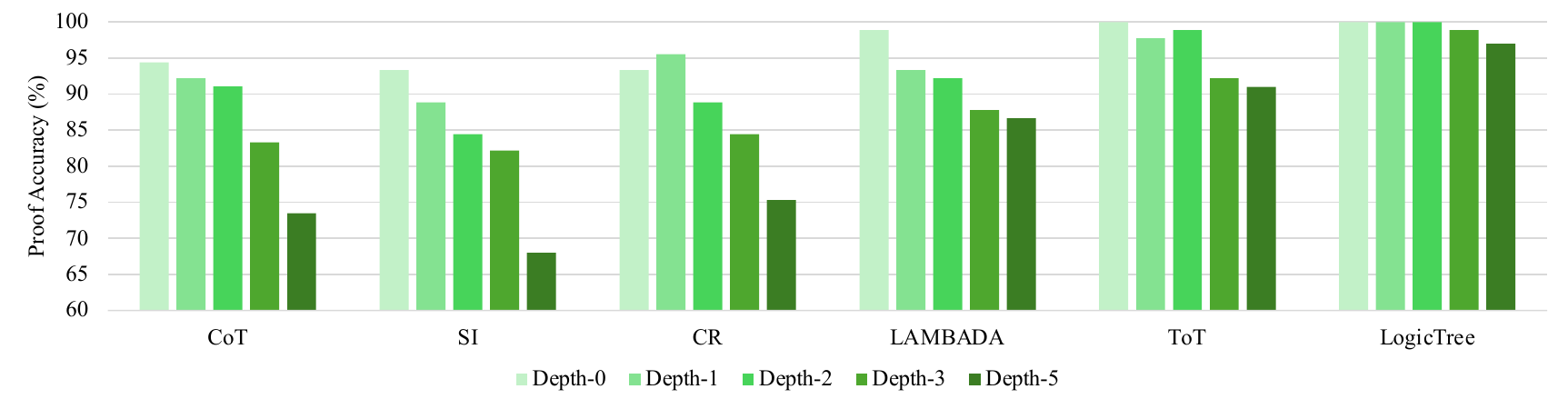}
\caption{Proof accuracy of different methods across different levels of task difficulty (measured by depth) in ProofWriter on GPT-4o. LogicTree consistently outperforms other methods at all depths, demonstrating superior robustness to problem difficulty.}
\label{fig7}
\end{figure*}
%%%%%%%%%%%%%%%%%%%%%%%%%%%%%%%%%%%%%%

%%%%%%%%%%%%%%%%%%%%%%%%%%%%%%%%%%%%%%
\begin{figure*}[t]
\centering
\includegraphics[width=1\textwidth]{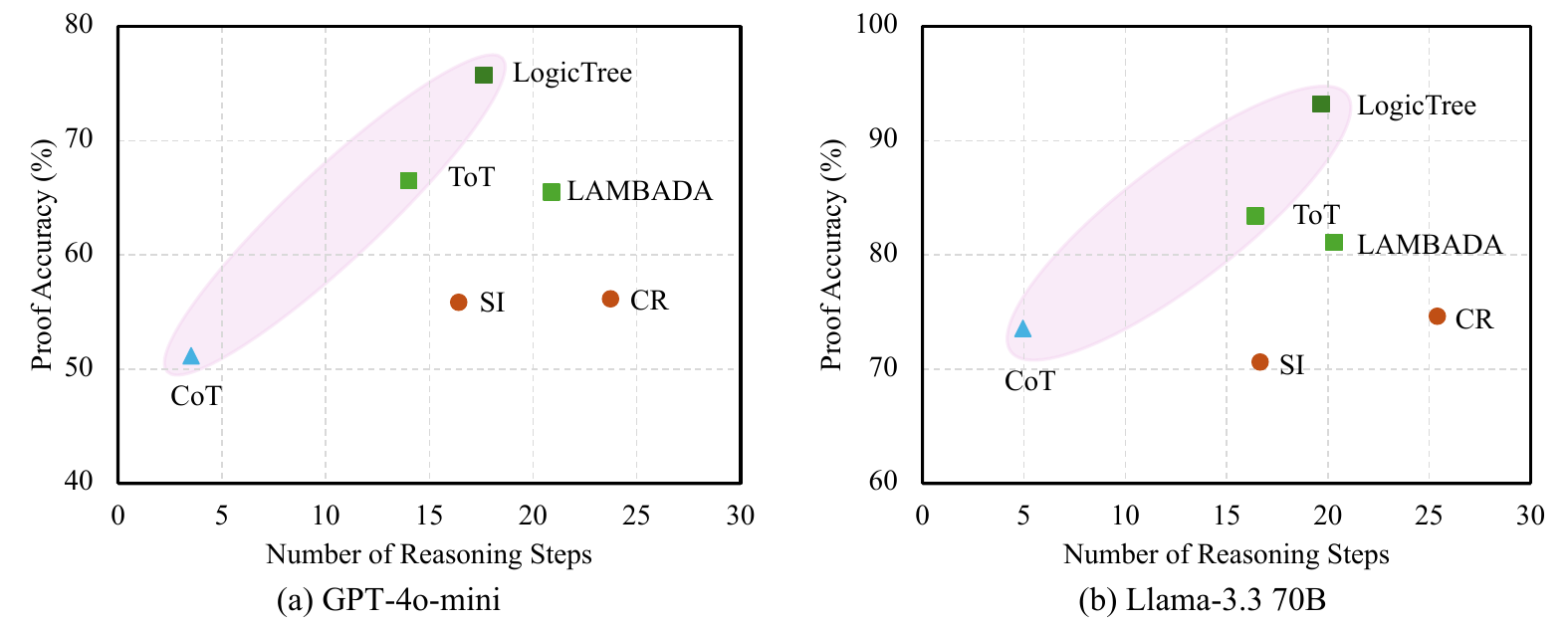}
\caption{Proof accuracy versus reasoning steps, averaged across five datasets for (a) GPT-4o-mini and (b) Llama-3.3 70B. The shaded area illustrates that our framework optimally scales inference-time computation to achieve higher proof accuracy.}
\label{fig8}
\end{figure*}
%%%%%%%%%%%%%%%%%%%%%%%%%%%%%%%%%%%%%%

%%%%%%%%%%%%%%%%%%%%%%%%%%%%%%%%%%%%%%
\begin{figure*}[t]
\centering
\includegraphics[width=1\textwidth]{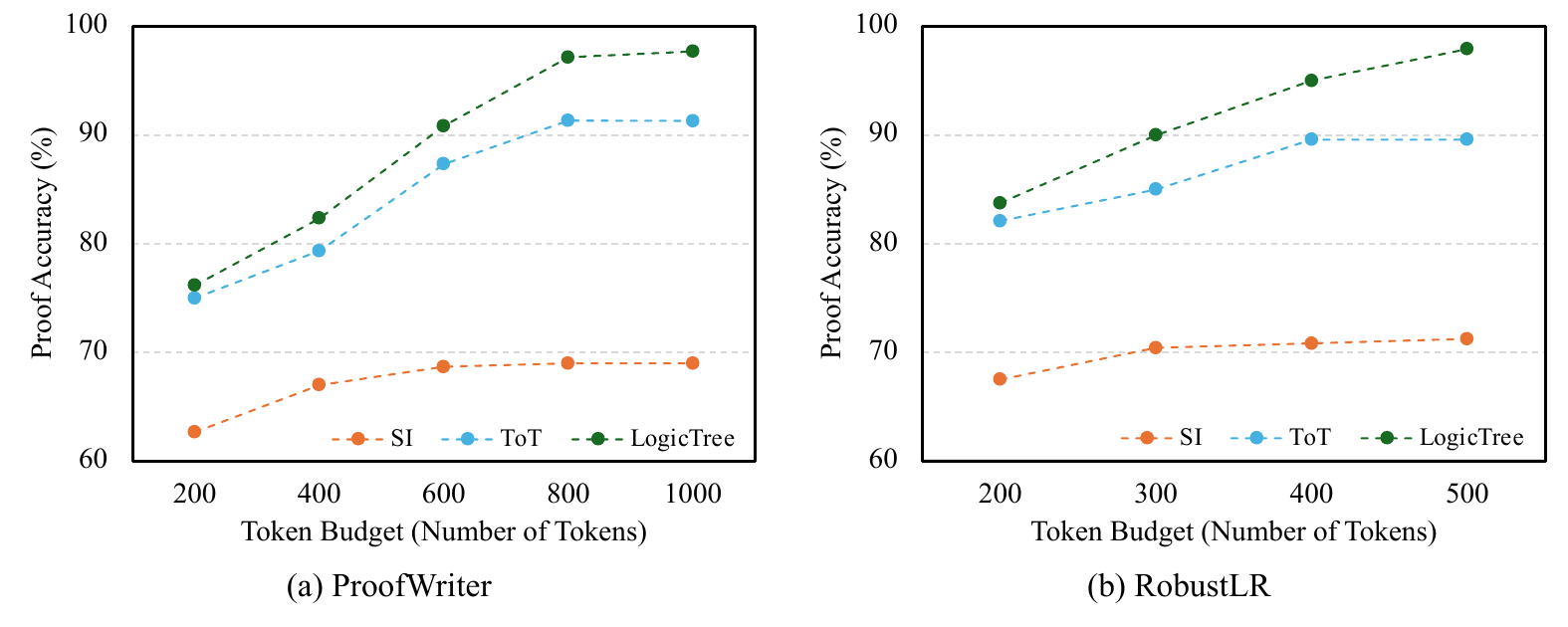}
\caption{Proof accuracy versus token budget for different methods (SI, ToT, LogicTree) on ProofWriter and RobustLR datasets, evaluated with GPT-4o.}
\label{token_budget}
\end{figure*}
%%%%%%%%%%%%%%%%%%%%%%%%%%%%%%%%%%%%%%

%%%%%%%%%%%%%%%%%%%%%%%%%%%%%%%%%%%%%%%%%%%%%%%%%
\begin{table*}[t]
\centering
\small
\renewcommand{\arraystretch}{1.4}
\setlength{\tabcolsep}{7pt}
  % \resizebox{1\textwidth}{!}{
\begin{tabular}{c c ccccc c}
    \toprule
    \multirow{2.5}{*}{\textbf{Model}} & \multirow{2.5}{*}{\textbf{Method}} & \multicolumn{5}{c}{\textbf{Dataset}} & \multirow{2.5}{*}{\textbf{Avg.\textsuperscript{\dag}}} \\
    \cmidrule(lr){3-7}
    & & \text{LogicNLI} & \text{ParaRules} & \text{PrOntoQA-OOD} & \text{ProofWriter} & \text{RobustLR} & \\
    \midrule

    %%%%%%%%
    \multirow{6}{*}{\centering \textbf{GPT-4o-mini}} 
      & CoT     & 3.4 & 3.0 & 3.7 & 3.5 & 4.8 & 3.5 \\
      
      \noalign{\vskip 1pt} % Adds small space before
      \cdashline{2-8}
      \noalign{\vskip 2pt} % Adds small space after

      & SI      & 16.6 & 18.9 & 8.6 & 18.5 & 11.7 & 16.4 \\
      & CR      & 25.7 & 23.1 & 22.1 & 23.3 & 26.7 & 23.8 \\
      & LAMBADA & 14.0 & 23.7 & 22.6 & 18.0 & 24.0 & 20.9 \\
      & ToT     & 36.2 & 5.1 & 21.7 & 17.0 & 8.5 & 14.0 \\
      
      \noalign{\vskip 1pt} % Adds small space before
      \cdashline{2-8}
      \noalign{\vskip 2pt} % Adds small space after
          
      \rowcolor{gray!20} % Light gray background
      & \textbf{LogicTree} & 43.2 & 6.8 & 28.2 & 21.2 & 11.1 & 17.6\\
    %%%%%%%%

    \midrule

    %%%%%%%%
    \multirow{6}{*}{\centering \textbf{GPT-4o}} 
      & CoT     & 5.2 & 3.5 & 3.8 & 4.3 & 6.6 & 4.4 \\
      
      \noalign{\vskip 1pt} % Adds small space before
      \cdashline{2-8}
      \noalign{\vskip 2pt} % Adds small space after

      & SI      & 19.6 & 17.0 & 10.7 & 19.4 & 14.6 & 17.0 \\
      & CR      & 24.1 & 25.1 & 21.8 & 25.7 & 23.4 & 24.6 \\
      & LAMBADA & 22.9 & 21.2 & 22.5 & 19.6 & 25.6 & 21.5 \\
      & ToT     & 32.9 & 6.4 & 23.1 & 20.2 & 10.3 & 15.7 \\
      
      \noalign{\vskip 1pt} % Adds small space before
      \cdashline{2-8}
      \noalign{\vskip 2pt} % Adds small space after
          
      \rowcolor{gray!20} % Light gray background
      & \textbf{LogicTree} & 45.6 & 9.5 & 32.2 & 23.5 & 13.8 & 20.3\\
    %%%%%%%%

    \midrule

    %%%%%%%%
    \multirow{6}{*}{\centering \textbf{ Llama-3.3 70B}} 
      & CoT     & 6.3 & 3.5 & 4.0 & 5.6 & 6.8 & 4.9 \\
      
      \noalign{\vskip 1pt} % Adds small space before
      \cdashline{2-8}
      \noalign{\vskip 2pt} % Adds small space after

      & SI      & 11.7 & 21.5 & 3.8 & 21.6 & 6.2 & 16.7 \\
      & CR      & 27.1 & 24.6 & 22.2 & 27.3 & 24.2 & 25.4 \\
      & LAMBADA & 19.6 & 25.2 & 17.9 & 14.0 & 26.4 & 20.3 \\
      & ToT     & 31.3 & 7.4 & 23.1 & 21.2 & 12.3 & 16.4 \\
      
      \noalign{\vskip 1pt} % Adds small space before
      \cdashline{2-8}
      \noalign{\vskip 2pt} % Adds small space after
          
      \rowcolor{gray!20} % Light gray background
      & \textbf{LogicTree} & 39.1 & 7.3 & 30.6 & 25.6 & 14.4 & 19.7\\
    %%%%%%%%

    \bottomrule
    \end{tabular}
  % }
  % \end{center}
  \caption{Average number of reasoning steps for different methods across five logical reasoning datasets on GPT-4o-mini, GPT-4o, and Llama-3.3 70B. Avg.\textsuperscript{\dag} is calculated as the total number of reasoning steps divided by the total number of examples across all five datasets.}
  \label{table_reason_step}
\end{table*}
%%%%%%%%%%%%%%%%%%%%%%%%%%%%%%%%%%%%%%%%%%%%%%%%%

%%%%%%%%%%%%%%%%%%%%%%%%%%%%%%%%%%%%%%
\begin{table*}[t]
\centering
\small
\renewcommand{\arraystretch}{1.4}
\setlength{\tabcolsep}{2.6pt}
% \resizebox{1\textwidth}{!}{
\begin{tabular}{c cc cc cc cc cc}
\toprule
\multirow{4}{*}{\textbf{Method}} & \multicolumn{10}{c}{\textbf{Dataset}} \\
\cmidrule(lr){2-11}
& \multicolumn{2}{c}{\text{LogicNLI}} 
& \multicolumn{2}{c}{\text{ParaRules}} 
& \multicolumn{2}{c}{\text{PrOntoQA-OOD}} 
& \multicolumn{2}{c}{\text{ProofWriter}} 
& \multicolumn{2}{c}{\text{RobustLR}} \\
\cmidrule(lr){2-3} \cmidrule(lr){4-5} \cmidrule(lr){6-7} \cmidrule(lr){8-9} \cmidrule(lr){10-11}
& Token \# & Time (min) 
& Token \# & Time (min) 
& Token \# & Time (min) 
& Token \# & Time (min) 
& Token \# & Time (min)  \\
\midrule
CoT      & 145.4 & 0.11 & 105.5 & 0.08 & 125.9 & 0.09 & 103.2 & 0.08 & 171.9 & 0.12  \\

% \noalign{\vskip 1pt} % Adds small space before
% \cdashline{1-11}
% \noalign{\vskip 2pt} % Adds small space after

SI       & 681.7 & 0.80 & 688.3 & 0.79 & 233.2 & 0.28 & 490.5 & 0.68 & 502.1 & 0.60 \\
CR       &  1231.6 & 1.43 & 871.6 & 1.04 & 490.4 & 0.62 & 816.7 & 0.94 & 800.7 & 0.97 \\
LAMBADA  & 1181.0 & 1.36 & 758.6 & 0.97 & 566.3 & 0.67 & 550.5 & 0.74 & 921.4 & 1.07 \\
ToT      & 1896.7 & 2.13 & 258.4 & 0.30 & 568.2 & 0.70 & 709.3 & 0.86 & 355.6 & 0.42 \\

% \noalign{\vskip 1pt} % Adds small space before
% \cdashline{1-11}
% \noalign{\vskip 2pt} % Adds small space after

\rowcolor{gray!8}
o1-mini  & 3253.3 & 1.11 & 464.5 & 0.19 & 943.0 & 0.39 & 953.1 & 0.42 & 534.8 & 0.24 \\
\rowcolor{gray!8}
o3-mini  & 2412.3 & 1.02 & 589.2 & 0.27 & 683.0 & 0.34 & 849.1 & 0.36 & 481.9 & 0.21 \\

% \noalign{\vskip 1pt} % Adds small space before
% \cdashline{1-11}
% \noalign{\vskip 2pt} % Adds small space after

\rowcolor{gray!20}
\textbf{LogicTree} & 3627.0 & 4.07 & 440.6 & 0.54 & 1082.0 & 1.47 & 738.4 & 0.89 & 493.9 & 0.61 \\
\bottomrule
\end{tabular}
% }
\caption{Average number of generated tokens and inference time (in minutes) for different methods across five logical reasoning datasets. CoT, SI, CR, LAMBADA, ToT, and LogicTree results are based on GPT-4o.}
\label{output_token_time}
\end{table*}
%%%%%%%%%%%%%%%%%%%%%%%%%%%%%%%%%%%%%%

\clearpage

%%%%%%%%%%%%%%%%%%%%%%%%%%%%%%%%%%%%%%
\begin{figure*}[t]
\centering
\includegraphics[width=1\textwidth]{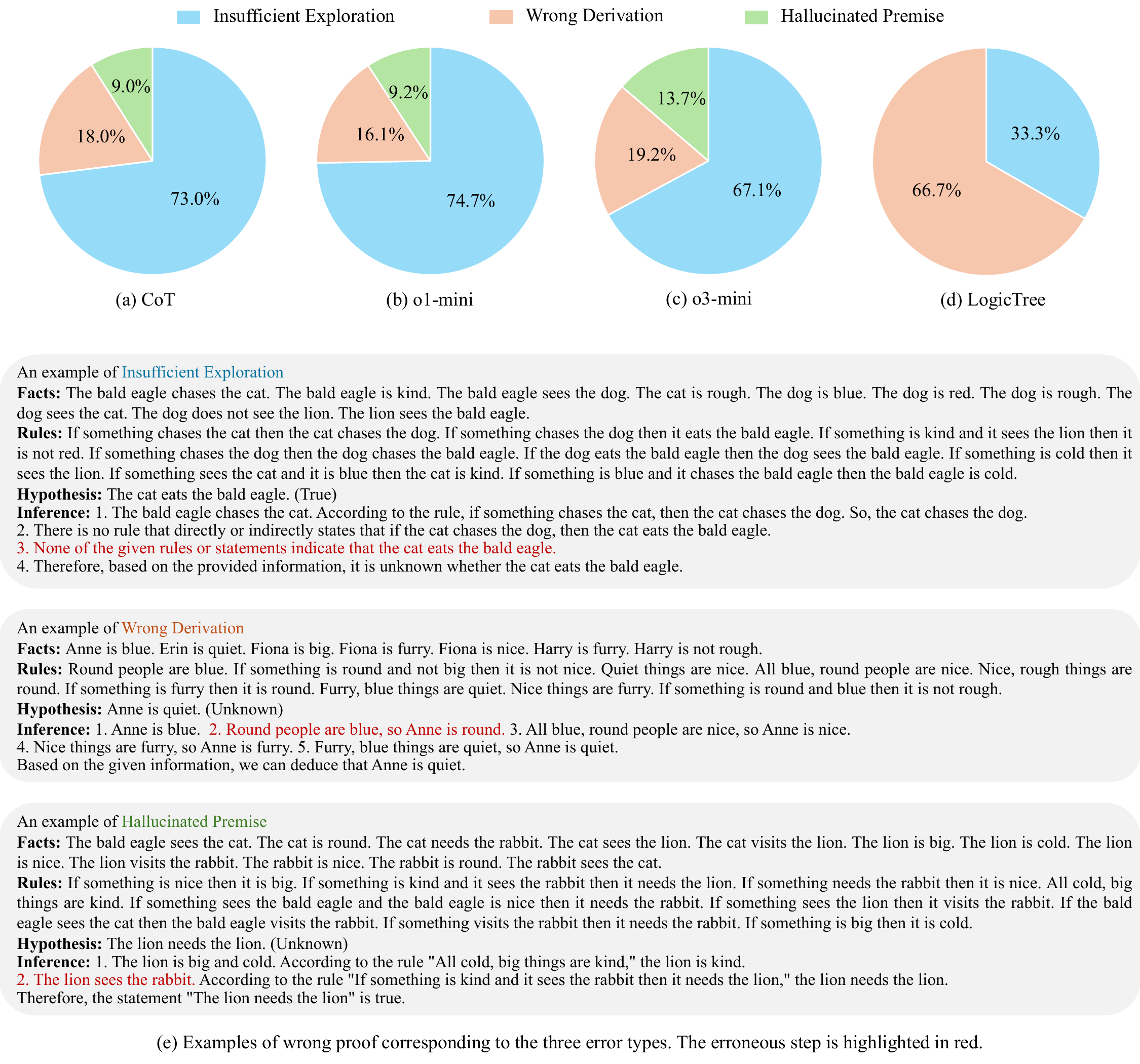}
\caption{Error analysis. (a–d) show the proportions of three error types for CoT (on GPT-4o), o1-mini, o3-mini, and LogicTree (on GPT-4o), respectively. (e) Examples of wrong proof corresponding to the three error types.}
\label{error_anals}
\end{figure*}
%%%%%%%%%%%%%%%%%%%%%%%%%%%%%%%%%%%%%%

%%%%%%%%%%%%%%%%%%%%%%%%%%%%%%%%%%%%%%
\begin{figure*}[t]
\centering
\includegraphics[width=1\textwidth]{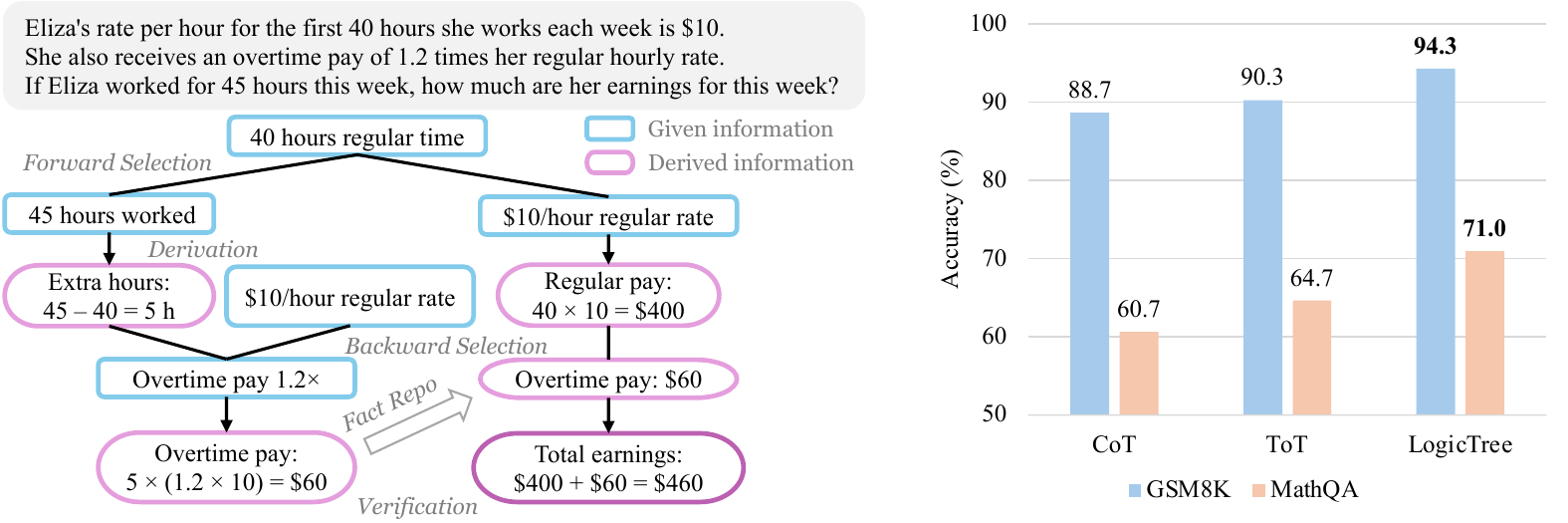}
\caption{LogicTree for mathematical reasoning. \textbf{Left}: Case study illustrating how LogicTree solves a math problem from GSM8K. \textbf{Right}: Accuracy comparison among CoT, ToT and LogicTree on GSM8K and MathQA.}
\label{math_demo}
\end{figure*}
%%%%%%%%%%%%%%%%%%%%%%%%%%%%%%%%%%%%%%
\clearpage

% --------------LogicNLI
\begin{figure*}[t]
\centering
\begin{tcolorbox}[
  width=\textwidth,
  % breakable,
  % enhanced,
  title=LogicNLI,
  colback=gray!5,
  colframe=nmgray!75!black,
  fontupper=\footnotesize\linespread{1}\selectfont
]
\noindent
\textbf{Premises (Facts + Rules):}\\
Carrick is filthy. Carrick is not financial. Galvin is grieving. Blake is filthy. Oscar is not relieved. Perry is not filthy. Blake is financial. Perry is relieved. Toby is financial. Perry is filthy. Oscar is not filthy. Toby is not filthy.

% \vspace{0.6em}
% \noindent
% \textbf{Rules:}\\
Someone who is filthy is always unlikely. It can be concluded that Carrick is not unlikely and Galvin is not filthy once knowing that Carrick is relieved and Perry is filthy. If there is at least one people who is both not relieved and filthy, then Blake is grieving. Someone being both filthy and not unlikely is equivalent to being relieved. If Blake is unlikely and Galvin is relieved, then Oscar is filthy. If Perry is relieved, then Carrick is not filthy, and vice versa. Carrick being not grieving or Toby being not filthy implies that Carrick is filthy. If Perry is not filthy or Carrick is not grieving, then Conway is not filthy. If there is at least one people who is not filthy, then Oscar is financial. Someone who is filthy is always both not filthy and not financial. If there is someone who is either not filthy or grieving, then Toby is not filthy. If there is someone who is both not grieving and filthy, then Blake is filthy.

\vspace{0.6em}
\noindent
\textbf{Hypothesis:}\\
Carrick is relieved.

\end{tcolorbox}
\caption{An example of LogicNLI~\citep{tian-etal-2021-diagnosing}.}
\label{logicnli}
\end{figure*}
% --------------LogicNLI

% --------------PrOntoQA-OOD
\begin{figure*}[t]
\centering
\begin{tcolorbox}[
  width=\textwidth,
  % breakable,
  % enhanced,
  title=PrOntoQA-OOD,
  colback=gray!5,
  colframe=nmgray!75!black,
  fontupper=\footnotesize\linespread{1}\selectfont
]
\noindent
\textbf{Premises (Facts + Rules):}\\
Rex is a tumpus. Rex is a vumpus. Rex is a lempus. Rex is a lempus. Rex is a wumpus. Rex is a jompus.

% \vspace{0.6em}
% \noindent
% \textbf{Rules:}\\
Zumpuses are grimpuses. Each dumpus is a gorpus. Everything that is a lempus, a wumpus, and a brimpus is a grimpus, a dumpus, and a zumpus. Each grimpus is an impus. Zumpuses are shumpuses. Grimpuses are gorpuses. Everything that is a lempus and a wumpus and a brimpus is a rompus. Everything that is a tumpus and a lempus and a vumpus is a gorpus. Grimpuses are yumpuses.

\vspace{0.6em}
\noindent
\textbf{Hypothesis:}\\
Rex is an impus.

\end{tcolorbox}
\caption{An example of PrOntoQA-OOD~\citep{saparov2023testing}.}
\label{prontoqa}
\end{figure*}
% --------------PrOntoQA-OOD

% --------------ProofWriter 
\begin{figure*}[t]
\centering
\begin{tcolorbox}[
  width=\textwidth,
  % breakable,
  % enhanced,
  title=ProofWriter,
  colback=gray!5,
  colframe=nmgray!75!black,
  fontupper=\footnotesize\linespread{1}\selectfont
]
\noindent
\textbf{Premises (Facts + Rules):}\\
The bald eagle chases the cow.
The bald eagle is kind.
The bald eagle is rough.
The bald eagle needs the rabbit.
The cow chases the rabbit.
The cow is cold.
The cow is green.
The cow is red.
The rabbit does not chase the bald eagle.
The rabbit chases the cow.
The rabbit does not eat the bald eagle.
The rabbit eats the cow.
The rabbit is cold.
The rabbit is green.
The squirrel eats the cow.
The squirrel does not eat the rabbit.

% \vspace{0.6em}
% \noindent
% \textbf{Rules:}\\
If something needs the bald eagle then the bald eagle chases the rabbit.
If the squirrel is rough and the squirrel is not kind then the squirrel is green.
If something chases the bald eagle then it needs the squirrel.
If something needs the rabbit then it chases the bald eagle.
If something chases the cow then the cow eats the bald eagle.
If something chases the bald eagle and it does not need the bald eagle then it is red.
If something needs the squirrel then the squirrel needs the rabbit.

\vspace{0.6em}
\noindent
\textbf{Hypothesis:}\\
The squirrel needs the rabbit.

\end{tcolorbox}
\caption{An example of ProofWriter~\citep{tafjord2020proofwriter}.}
\label{proofwriter}
\end{figure*}
% --------------ProofWriter 

% --------------RobustLR
\begin{figure*}[t]
\centering
\begin{tcolorbox}[
  width=\textwidth,
  % breakable,
  % enhanced,
  title=RobustLR,
  colback=gray!5,
  colframe=nmgray!75!black,
  fontupper=\footnotesize\linespread{1}\selectfont
]
\noindent
\textbf{Premises (Facts + Rules):}\\
Fiona is not Bob's mother.
Harry is Charlie's son.

% \vspace{0.6em}
% \noindent
% \textbf{Rules:}\\
The father of Dave is Bob if Gary is not green.
If Fiona is not Bob's mother then Charlie is not Dave's aunt.
If Fiona is not Bob's son then Charlie is the aunt of Dave.
If Bob is rough then Bob is Dave's daughter.
Fiona is not the son of Bob if Bob is rough.
Dave is not kind if Fiona is not the son of Bob.
If The son of Bob is not Fiona then Harry is not white.
If Fiona is Harry's grandfather then Harry is white.
If Bob is rough then The grandfather of Harry is not Fiona.
Anne is not furry if The aunt of Dave is not Charlie.
Bob is not Dave's father if Bob is rough.
Gary is green if Bob is rough.
The husband of Dave is not Anne if The son of Bob is not Fiona.
If Bob is not the daughter of Dave then Gary is not green.
Fiona is the grandfather of Harry if The mother of Bob is not Fiona.
If Anne is not the husband of Dave then Anne is furry.
If Harry is white and The son of Charlie is Harry then The daughter of Dave is Bob.

\vspace{0.6em}
\noindent
\textbf{Hypothesis:}\\
The daughter of Dave is not Bob.

\end{tcolorbox}
\caption{An example of RobustLR~\citep{sanyal-etal-2022-robustlr}.}
\label{RobustLR}
\end{figure*}
% --------------RobustLR

% --------------ParaRules
\begin{figure*}[t]
\centering
\begin{tcolorbox}[
  width=\textwidth,
  % breakable,
  % enhanced,
  title=ParaRules,
  colback=gray!5,
  colframe=nmgray!75!black,
  fontupper=\footnotesize\linespread{1}\selectfont
]
\noindent
\textbf{Premises (Facts + Rules):}\\
Bob is a cold and round man who has red and green skin. Charlie is a kind person and he is also often cold. That guy Eric sure is nice. Harry is a really nice guy with a big round body, usually wearing red.

% \vspace{0.6em}
% \noindent
% \textbf{Rules:}\\
People who are round and red tend to be rough. If a person acts cold yet nice and green, they will be kind. If you meet someone with rough skin who is cold from being outside, you'll notice they are nice. Every time you meet someone kind and nice, they'll be green, too. Big people with red hair are cold because they cannot find coats that fit. It's a certainty that any green, big and kind individual is going to be nice. A big round young person is often blue.

\vspace{0.6em}
\noindent
\textbf{Hypothesis:}\\
Bob is nice.

\end{tcolorbox}
\caption{An example of ParaRules~\citep{ijcai2020p0537}.}
\label{ParaRules}
\end{figure*}
% --------------ParaRules

% -------------- o1/o3 prompt
\begin{figure*}[t]
\centering
\begin{tcolorbox}[
  width=\textwidth,
  % breakable,
  % enhanced,
  title={Prompts for o1-mini \& o3-mini},
  colback=gray!5,
  colframe=nmgray!75!black,
  fontupper=\footnotesize\linespread{1}\selectfont
]
\noindent
\textbf{Instructions:}\\
You will do logic reasoning tasks. You will be given a set of premises and a hypothesis. You need to answer if the hypothesis is *True* or *False* or *Unknown* based on the premises.\\
(The last sentence in the response should be in the format of "Therefore, the hypothesis is True / False / Unknown.")\\
\rule{2cm}{0.4pt}

\vspace{0.6em}
\noindent
\textbf{Query:}\\
Premise:\\
\textcolor{gray}{\texttt{query\_premise}}\\[0.3em]
Hypothesis:\\
\textcolor{gray}{\texttt{query\_hypothesis}}\\[0.3em]
Reasoning:\\
\textcolor{gray}{\texttt{LLM\_output}}

\end{tcolorbox}
\caption{The prompts for reasoning with OpenAI’s o1-mini and o3-mini model. Following OpenAI’s guidance~\citep{openai2024prompt}, we adopt zero-shot prompting and keep the prompts simple and direct.}
\label{o1_o3_prompt}
\end{figure*}

% -------------- CoT prompt
\begin{figure*}[t]
\centering
\begin{tcolorbox}[
  width=\textwidth,
  % breakable,
  % enhanced,
  title=Prompts for Chain-of-Thought,
  colback=gray!5,
  colframe=nmgray!75!black,
  fontupper=\footnotesize\linespread{1}\selectfont
]
\noindent
\textbf{Instructions:}\\
Suppose you are one of the greatest AI scientists, logicians. Given some context as premise, the task is to answer if a logical reasoning question (hypothesis) is *True* or *False* or *Unknown*.\\
Definitions:\\
- True: The hypothesis can be logically derived from the premises.\\
- False: Only the negation of the hypothesis can be logically derived from the premises.\\
- Unknown: Neither the hypothesis nor its negation can be logically derived from the premises.\\
Let us think step by step.

\vspace{0.6em}
\noindent
\textbf{Demonstrations:}\\
\#\#\#\#\# \texttt{Example\_1} \\
\textit{\# Example's hypothesis is *True*}\\
......\\[0.3em]
\#\#\#\#\# \texttt{Example\_2} \\
\textit{\# Example's hypothesis is *False*}\\
Premise:\\
\textcolor{gray}{The lion is cold.\\
The lion is nice.\\
The lion likes the squirrel.\\
The squirrel is round.\\
The squirrel sees the lion.\\
The squirrel visits the cow.\\
The tiger likes the cow.\\
The tiger likes the squirrel.\\
If something is cold then it visits the tiger.\\
If something visits the tiger then it is nice.\\
If something sees the tiger and it is young then it is blue.\\
If something is nice then it sees the tiger.\\
If something likes the squirrel and it likes the cow, then it visits the tiger.\\
If something is nice and it sees the tiger then it is young.\\
If the cow is cold and the cow visits the lion then the lion sees the squirrel.}\\[0.3em]
Question:\\
\textcolor{gray}{Based on the above information, is the following statement true, false, or unknown? The tiger is not young.}\\[0.3em]
Reasoning: \textit{\# Expected results}\\
\textcolor{gray}{1. The tiger likes the cow. The tiger likes the squirrel. If something likes the squirrel and it likes the cow, then it visits the tiger. So the tiger visits the tiger. \\
2. If something visits the tiger then it is nice. So the tiger is nice.\\
3. If something is nice then it sees the tiger. So the tiger sees the tiger.\\
4. If something is nice and it sees the tiger then it is young. So the tiger is young.\\
5. It contradicts "The tiger is not young." in the given question. So the answer is False.}\\[0.3em]
\#\#\#\#\# \texttt{Example\_3} \\
\textit{\# Example's hypothesis is *Unknown*}\\
......\\
\rule{2cm}{0.4pt}

\vspace{0.6em}
\noindent
\textbf{Query:}\\
Premise:\\
\textcolor{gray}{\texttt{query\_premise}}\\[0.3em]
Question:\\
\textcolor{gray}{\texttt{query\_hypothesis}}\\[0.3em]
Reasoning:\\
\textcolor{gray}{\texttt{LLM\_output}}

\end{tcolorbox}
\caption{The prompts for chain-of-thought reasoning. We number the reasoning steps in demonstrations to make the step counting explicit in LLM’s output. Demonstrations with hypotheses labeled as \textit{True} and \textit{Unknown} are omitted for brevity.}
\label{cot_prompt}
\end{figure*}

% -------------- forward selection prompt
\begin{figure*}[t]
\centering
\begin{tcolorbox}[
  width=\textwidth,
  % breakable,
  % enhanced,
  title=Prompts for Forward Selection Module,
  colback=gray!5,
  colframe=nmgray!75!black,
  fontupper=\footnotesize\linespread{1}\selectfont
]
\noindent
\textbf{Instructions:}\\
Imagine you are one of the greatest AI scientists. You are given **a fact** and **a list of rules** (each rule being a premise with condition(s)). Your task is to evaluate each rule in the list and select those that meet *any* of the following requirements:\\
- Full Condition Match: The fact fully and directly satisfies all condition(s) of the rule, allowing a valid derivation to obtain a new proposition.\\
- Partial Condition Match: The fact directly satisfies some, but not all, conditions of the rule. This means that additional fact(s) would be required to make a full derivation and obtain a new proposition.\\
If no rule is selected, return **None**.

\vspace{0.6em}
\noindent
\textbf{Demonstrations:}\\
\#\#\#\#\# \texttt{Example\_1}\\
The given fact:\\
\textcolor{gray}{Bob is red.}\\[0.3em]
The given list of rules:\\
\textcolor{gray}{All red, round people are quiet.\\
Red people are young.\\
If someone is round and smart then they are not red.\\
All white people are red.\\
Quiet people are green.\\
If someone is red and not white then they are not green.\\
If someone likes the dog and they are red then they are blue.}\\[0.3em]
Let's go through each rule from the given list of rules and think step by step.\\
The selected rules (partial or full condition directly matched) are: \textit{\# Expected results}\\
\textcolor{gray}{All red, round people are quiet.\\
Red people are young.\\
If someone is red and not white then they are not green.\\
If someone likes the dog and they are red then they are blue.}\\[0.3em]
% -----------------------------
\#\#\#\#\# \texttt{Example\_2}\\
The given fact:\\
\textcolor{gray}{Anne is quiet.}\\[0.3em]
The given list of rules:\\
\textcolor{gray}{If something is furry and not blue then it is nice.\\
If Anne is furry then Anne is nice.\\
Smart, furry things are round.}\\[0.3em]
Let's go through each rule from the given list of rules and think step by step.\\
The selected rules (partial or full condition directly matched) are: \textit{\# Expected results}\\
\textcolor{gray}{None}\\
\rule{2cm}{0.4pt}

\vspace{0.6em}
\noindent
\textbf{Query:}\\
The given fact:\\
\textcolor{gray}{\texttt{query\_given\_fact}}\\[0.3em]
The given list of rules:\\
\textcolor{gray}{\texttt{query\_given\_list\_of\_rules}}\\[0.3em]
Let's go through each rule from the given list of rules and think step by step.\\
The selected rules (partial or full condition directly matched) are:\\
\textcolor{gray}{\texttt{LLM\_output}}

\end{tcolorbox}
\caption{The prompts for Forward Selection Module for \textit{rule selection} in LogicTree.}
\label{forward_prompt}
\end{figure*}

% -------------- backward selection prompt
\begin{figure*}[t]
\centering
\begin{tcolorbox}[
  width=\textwidth,
  % breakable,
  % enhanced,
  title=Prompts for Backward Selection Module,
  colback=gray!5,
  colframe=nmgray!75!black,
  fontupper=\footnotesize\linespread{1}\selectfont
]
\noindent
\textbf{Instructions:}\\
Suppose you are one of the greatest AI scientists, logicians. Given a specific fact, a rule, and a repository of facts, your task is to identify the missing fact(s) required to fully satisfy the rule's conditions and check if the missing fact(s) exist in the fact repository.\\
- The given one specific fact already satisfies one of the rule's conditions. Identify the missing fact(s) needed to fully satisfy the rule.\\
- Automatically adapt pronouns (e.g., 'they', 'something', 'someone') to the correct subject based on the context of the given rule and the given fact.\\
- Check if the missing fact(s) are present in the fact repository.\\
\hspace*{1em}- If the missing fact(s) are present in the fact repository, return **True** along with the identified missing fact(s).\\
\hspace*{1em}- Otherwise,  return **False**.

\vspace{0.6em}
\noindent
\textbf{Demonstrations:}\\
\#\#\#\#\# \texttt{Example\_1}\\
The given one specific fact:\\
\textcolor{gray}{The cat likes the rabbit.}\\[0.3em]
The given rule:\\
\textcolor{gray}{If someone is cold and they like the rabbit then the rabbit likes the cat.}\\[0.3em]
The given fact repository:\\
\textcolor{gray}{The cat eats the bear.\\
The cat is cold.\\
The cat is kind.\\
The cat likes the rabbit.\\
The rabbit likes the tiger.\\
The tiger likes the bear.\\
The tiger visits the cat.}\\[0.3em]
Let's go through each condition of the given rule. First identify the missing fact(s) needed to fully satisfy the rule.\\
Then check if the missing fact(s) are present in the fact repository: \textit{\# Expected results}\\
\textcolor{gray}{The cat is cold.\\ True. The identified missing fact(s) in the fact repository: The cat is cold.}\\[0.3em]
% -----------------------------
\#\#\#\#\# \texttt{Example\_2}\\
The given one specific fact:\\
\textcolor{gray}{The rabbit likes the squirrel.}\\[0.3em]
The given rule:\\
\textcolor{gray}{If someone likes the squirrel and the squirrel sees the cow then they are red.}\\[0.3em]
The given fact repository:\\
\textcolor{gray}{The cow likes the rabbit.\\
The cow needs the mouse.\\
The mouse likes the squirrel.\\
The rabbit needs the cow.\\
The rabbit sees the cow.\\
The squirrel is nice.\\
The squirrel needs the cow.\\
The rabbit likes the squirrel.}\\[0.3em]
Let's go through each condition of the given rule. First identify the missing fact(s) needed to fully satisfy the rule.\\
Then check if the missing fact(s) are present in the fact repository: \textit{\# Expected results}\\
\textcolor{gray}{The squirrel sees the cow.\\ False}\\
\rule{2cm}{0.4pt}

\vspace{0.6em}
\noindent
\textbf{Query:}\\
The given one specific fact:\\
\textcolor{gray}{\texttt{query\_given\_fact}}\\[0.3em]
The given rule:\\
\textcolor{gray}{\texttt{query\_given\_rule}}\\[0.3em]
The given fact repository:\\
\textcolor{gray}{\texttt{query\_given\_fact\_repo}}\\[0.3em]
Let's go through each condition of the given rule. First identify the missing fact(s) needed to fully satisfy the rule.\\
Then check if the missing fact(s) are present in the fact repository:\\
\textcolor{gray}{\texttt{LLM\_output}}

\end{tcolorbox}
\caption{The prompts for Backward Selection Module for \textit{fact selection} in LogicTree.}
\label{backward_prompt}
\end{figure*}

% -------------- derivation prompt
\begin{figure*}[t]
\centering
\begin{tcolorbox}[
  width=\textwidth,
  % breakable,
  % enhanced,
  title=Prompts for Derivation Module,
  colback=gray!5,
  colframe=nmgray!75!black,
  fontupper=\footnotesize\linespread{1}\selectfont
]
\noindent
\textbf{Instructions:}\\
Suppose you are one of the greatest AI scientists, logicians. Your task is to derive a new **Proposition** based on a given **rule** and some **fact(s)**. \\
Follow these instructions carefully:\\
1. Ensure that the **Proposition**:\\
    \hspace*{1em} - Must be a valid logical derivation from the provided **rule** and **fact(s)**.\\
    \hspace*{1em} - Must not duplicate any of the provided **fact(s)**.\\
    \hspace*{1em} - Must not include any information not directly derived from the provided information.\\
    \hspace*{1em} - Automatically adapt pronouns (e.g., 'they', 'something', 'someone') to the correct subject based on the context.\\
2. Do not apply the rule unless all conditions of the rule are met.\\
3. If no new **Proposition** can be derived, return **None**, and classify the reason into one of the following categories:\\
    \hspace*{1em} - A. **Partial Information Met**: The given fact(s) meet some but not all conditions of the given rule.\\
    \hspace*{1em} - B. **No Information Met**: The given fact(s) do not meet any conditions of the given rule.

\vspace{0.6em}
\noindent
\textbf{Demonstrations:}\\
\#\#\#\#\# \texttt{Example\_1}\\
The given fact(s):\\
\textcolor{gray}{Erin is tall. Erin is cold.}\\[0.3em]
The given rule:\\
\textcolor{gray}{Cold, tall people are not furry.}\\[0.3em]
The derived proposition is: \textit{\# Expected results}\\
\textcolor{gray}{Erin is not furry.}\\[0.3em]
% -----------------------------
\#\#\#\#\# \texttt{Example\_2}\\
The given fact(s):\\
\textcolor{gray}{Bob is round.}\\[0.3em]
The given rule:\\
\textcolor{gray}{If someone is round and smart then they are not red.}\\[0.3em]
The derived proposition is: \textit{\# Expected results (pseudo dead-end)}\\
\textcolor{gray}{None\\ Reason: A. **Partial Information Met**: The given fact(s) meet some but not all conditions of the given rule.}\\[0.3em]
% -----------------------------
\#\#\#\#\# \texttt{Example\_3}\\
The given fact(s):\\
\textcolor{gray}{Alice is happy.}\\[0.3em]
The given rule:\\
\textcolor{gray}{If Alice is sad and red, she is quiet.}\\[0.3em]
The derived proposition is: \textit{\# Expected results (dead end)}\\
\textcolor{gray}{None\\ Reason: B. **No Information Met**: The given fact(s) do not meet any conditions of the given rule.}\\
\rule{2cm}{0.4pt}

\vspace{0.6em}
\noindent
\textbf{Query:}\\
The given fact(s):\\
\textcolor{gray}{\texttt{query\_given\_facts}}\\[0.3em]
The given rule:\\
\textcolor{gray}{\texttt{query\_given\_rule}}\\[0.3em]
The derived proposition is:\\
\textcolor{gray}{\texttt{LLM\_output}}

\end{tcolorbox}
\caption{The prompts for Derivation Module in LogicTree.}
\label{derive_prompt}
\end{figure*}

% -------------- verification prompt
\begin{figure*}[t]
\centering
\begin{tcolorbox}[
  width=\textwidth,
  % breakable,
  % enhanced,
  title=Prompts for Verification Module,
  colback=gray!5,
  colframe=nmgray!75!black,
  fontupper=\footnotesize\linespread{1}\selectfont
]
\noindent
\textbf{Instructions:}\\
Suppose you are one of the greatest AI scientists, logicians. Your task is to verify the relationship between a given **Proposition** and a **Conclusion**. There are three possibilities:\\
1. **Same:** The **Proposition** is directly equivalent to the **Conclusion**, meaning both the subject and the predicate (attributes) are the same.\\
2. **Opposite:** The **Proposition** directly contradicts the **Conclusion**. The subjects are the same, but the predicates (attributes) are in direct opposition, such as 'predicate' versus 'not predicate'.\\
3. **Indeterminate:** Neither **Same** nor **Opposite**. The **Proposition** and the **Conclusion** either have different predicates (attributes) or there is no clear relationship between them.

\vspace{0.6em}
\noindent
\textbf{Demonstrations:}\\
\#\#\#\#\# \texttt{Example\_1}\\
Proposition:\\
\textcolor{gray}{Erin is not round.}\\[0.3em]
Conclusion:\\
\textcolor{gray}{Erin is not green.}\\[0.3em]
Verify the relationship between the given Proposition and the Conclusion: \textit{\# Expected results}\\
\textcolor{gray}{Indeterminate}\\[0.3em]
% -----------------------------
\#\#\#\#\# \texttt{Example\_2}\\
Proposition:\\
\textcolor{gray}{The rabbit is cold.}\\[0.3em]
Conclusion:\\
\textcolor{gray}{The rabbit is cold.}\\[0.3em]
Verify the relationship between the given Proposition and the Conclusion: \textit{\# Expected results}\\
\textcolor{gray}{Same}\\[0.3em]
% -----------------------------
\#\#\#\#\# \texttt{Example\_3}\\
Proposition:\\
\textcolor{gray}{The tiger is not young.}\\[0.3em]
Conclusion:\\
\textcolor{gray}{The tiger is young.}\\[0.3em]
Verify the relationship between the given Proposition and the Conclusion: \textit{\# Expected results}\\
\textcolor{gray}{Opposite}\\
\rule{2cm}{0.4pt}

\vspace{0.6em}
\noindent
\textbf{Query:}\\
Proposition:\\
\textcolor{gray}{\texttt{query\_proposition}}\\[0.3em]
Conclusion:\\
\textcolor{gray}{\texttt{query\_conclusion}}\\[0.3em]
Verify the relationship between the given Proposition and the Conclusion:\\
\textcolor{gray}{\texttt{LLM\_output}}

\end{tcolorbox}
\caption{The prompts for Verification Module in LogicTree.}
\label{verify_prompt}
\end{figure*}

% -------------- fact rank prompt
\begin{figure*}[t]
\centering
\begin{tcolorbox}[
  width=\textwidth,
  % breakable,
  % enhanced,
  title=Prompts for Fact Ranking Module (ablation study),
  colback=gray!5,
  colframe=nmgray!75!black,
  fontupper=\footnotesize\linespread{1}\selectfont
]
\noindent
\textbf{Instructions:}\\
Imagine you are one of the greatest AI scientists, logicians. You are given a logic reasoning question that involves: a list of facts, a list of rules, a hypothesis to be verified.\\
Your task is to plan and prioritize the reasoning path:\\
- Sort the given **facts** based on their likelihood of being the starting point in the correct reasoning path to verify the hypothesis.\\
- The first fact in the sorted list should have the highest probability of being the right starting point, and the last fact should have the lowest probability.

\vspace{0.6em}
\noindent
\textbf{Demonstrations:}\\
\#\#\#\#\# \texttt{Example\_1}\\
The given list of facts:\\
\textcolor{gray}{Bob is young.\\
Dave is blue.\\
Erin is blue.\\
Fiona is blue.\\
Fiona is kind.\\
Fiona is quiet.\\
Fiona is white.}\\[0.3em]
The given list of rules:\\
\textcolor{gray}{If someone is kind then they are white.\\
Young people are quiet.\\
If someone is kind and white then they are blue.\\
All quiet, kind people are white.\\
If someone is quiet then they are kind.\\
If someone is white then they are young.\\
All blue, kind people are green.}\\[0.3em]
The hypothesis to be verified:\\
\textcolor{gray}{Fiona is not green.}\\[0.3em]
Let's sort the given **facts** based on their likelihood of being the starting point in the correct reasoning path.\\
The sorted facts are (each fact in a new line): \textit{\# Expected results}\\
\textcolor{gray}{Fiona is blue.\\
Fiona is kind.\\
Fiona is quiet.\\
Fiona is white.\\
Bob is young.\\
Dave is blue.\\
Erin is blue.}\\[0.3em]
% -----------------------------
\#\#\#\#\# \texttt{Example\_2}\\
......\\
\rule{2cm}{0.4pt}

\vspace{0.6em}
\noindent
\textbf{Query:}\\
The given list of facts:\\
\textcolor{gray}{\texttt{query\_fact\_list}}\\[0.3em]
The given list of rules:\\
\textcolor{gray}{\texttt{query\_rule\_list}}\\[0.3em]
The hypothesis to be verified:\\
\textcolor{gray}{\texttt{query\_hypothesis}}\\[0.3em]
Let's sort the given **facts** based on their likelihood of being the starting point in the correct reasoning path.\\
The sorted facts are (each fact in a new line):\\
\textcolor{gray}{\texttt{LLM\_output}}

\end{tcolorbox}
\caption{Prompts for Fact Ranking Module, used in the ablation study reported in Table~\ref{table2}.}
\label{fact_rank_prompt}
\end{figure*}

% -------------- rule rank prompt
\begin{figure*}[t]
\centering
\begin{tcolorbox}[
  width=\textwidth,
  % breakable,
  % enhanced,
  title=Prompts for Rule Ranking Module (ablation study),
  colback=gray!5,
  colframe=nmgray!75!black,
  fontupper=\footnotesize\linespread{1}\selectfont
]
\noindent
\textbf{Instructions:}\\
Imagine you are one of the greatest AI scientists, logicians. You are given a logic reasoning question that involves: a list of facts, a list of rules, a hypothesis to be verified.\\
Additionally, you are provided with a set of **selected rules**, which serve as potential intermediate steps in the reasoning process.\\
Your task is to plan and prioritize the reasoning path:\\
- Sort the **selected rules** based on their likelihood of being part of the correct reasoning path.\\
- The first rule in the sorted list should have the highest probability of being in the correct reasoning path, and the last rule should have the lowest probability.

\vspace{0.6em}
\noindent
\textbf{Demonstrations:}\\
\#\#\#\#\# \texttt{Example\_1}\\
The given list of facts:\\
\textcolor{gray}{Bob is young.\\
Dave is blue.\\
Erin is blue.\\
Fiona is blue.\\
Fiona is kind.\\
Fiona is quiet.\\
Fiona is white.}\\[0.3em]
The given list of rules:\\
\textcolor{gray}{If someone is kind then they are white.\\
Young people are quiet.\\
If someone is kind and white then they are blue.\\
All quiet, kind people are white.\\
If someone is quiet then they are kind.\\
If someone is white then they are young.\\
All blue, kind people are green.}\\[0.3em]
The hypothesis to be verified:\\
\textcolor{gray}{Fiona is not green.}\\[0.3em]
The given set of **selected rules**:\\
\textcolor{gray}{If someone is kind then they are white.\\
If someone is kind and white then they are blue.\\
All quiet, kind people are white.\\
All blue, kind people are green.}\\[0.3em]
Let's sort the given **selected rules** based on their likelihood of being part of the correct reasoning path.\\
The sorted rules are (each rule in a new line): \textit{\# Expected results}\\
\textcolor{gray}{All blue, kind people are green.\\
If someone is kind then they are white.\\
If someone is kind and white then they are blue.\\
All quiet, kind people are white.}\\[0.3em]
% -----------------------------
\#\#\#\#\# \texttt{Example\_2}\\
......\\
\rule{2cm}{0.4pt}

\vspace{0.6em}
\noindent
\textbf{Query:}\\
The given list of facts:\\
\textcolor{gray}{\texttt{query\_fact\_list}}\\[0.3em]
The given list of rules:\\
\textcolor{gray}{\texttt{query\_rule\_list}}\\[0.3em]
The hypothesis to be verified:\\
\textcolor{gray}{\texttt{query\_hypothesis}}\\[0.3em]
The given set of **selected rules**:\\
\textcolor{gray}{\texttt{query\_selected\_rules}}\\[0.3em]
Let's sort the given **selected rules** based on their likelihood of being part of the correct reasoning path.\\
The sorted rules are (each rule in a new line):\\
\textcolor{gray}{\texttt{LLM\_output}}

\end{tcolorbox}
\caption{Prompts for Rule Ranking Module, used in the ablation study reported in Table~\ref{table2}.}
\label{rule_rank_prompt}
\end{figure*}

\end{document}